\documentclass[10pt,twocolumn,letterpaper]{article}

\usepackage[pagenumbers]{cvpr} 

\definecolor{cvprblue}{rgb}{0.21,0.49,0.74}
\usepackage[pagebackref,breaklinks,colorlinks,allcolors=cvprblue]{hyperref}
\usepackage{url}
\usepackage{graphicx}  
\usepackage{subcaption} 
\usepackage{enumitem}
\usepackage{booktabs}
\usepackage{array}
\usepackage{float}
\usepackage{tikz}
\usepackage{algorithm}
\usepackage{listings}
\usepackage{algpseudocode}
\usepackage{amsmath}
\usepackage{bm}

\usepackage{lscape}
\usepackage{multirow}
\usepackage{etoc}
\usepackage[absolute,overlay]{textpos}
\usepackage{graphicx}  
\usepackage{wrapfig}   
\etocsettagdepth{mtappendix}{none}
\etocdepthtag.toc{mtchapter}

\def\paperID{762} 
\def\confName{CVPR}
\def\confYear{2025}

\title{Enhancing Perception Capabilities of \\ Multimodal LLMs with Training-Free Fusion}

\author{
    Zhuokun Chen\textsuperscript{\rm 1 \rm 2}\thanks{Email: caesard216@gmail.com} ~~ 
    Jinwu Hu\textsuperscript{\rm 1 \rm 2} ~~
    Zeshuai Deng\textsuperscript{\rm 1} ~~
    Yufeng Wang\textsuperscript{\rm 1} \\
    Bohan Zhuang\textsuperscript{\rm 3}\footnotemark[2] ~
    Mingkui Tan\textsuperscript{\rm 1 \rm 2 \rm 4}\thanks{Corresponding authors.} ~ \\
    \textsuperscript{\scriptsize{\rm 1}}\small{South China University of Technology,}
    \textsuperscript{\rm 2}\small{Pazhou Lab,}
    \textsuperscript{\rm 3}\small{ZIP Lab, Zhejiang University,} \\
    \textsuperscript{\rm 4}\small{Key Laboratory of Big Data and Intelligent Robot, Ministry of Education}\\
}

\newcommand{\methodname}{VisionFuse}

\begin{document}
\maketitle
\begin{abstract}
Multimodal LLMs (MLLMs) equip language models with visual capabilities by aligning vision encoders with language models. 
Existing methods to enhance the visual perception of MLLMs often involve designing more powerful vision encoders, which requires exploring a vast design space and re-aligning each potential encoder with the language model, resulting in prohibitively high training costs.
In this paper, we introduce VisionFuse, a novel integration framework that efficiently utilizes multiple vision encoders from off-the-shelf MLLMs to enhance visual perception without requiring additional training.
Our approach is motivated by the observation that different MLLMs tend to focus on distinct regions given the same query and image. Moreover, we find that the feature distributions of vision encoders within an MLLM family, a group of MLLMs sharing the same pretrained LLM, are highly aligned.
Building on these insights, VisionFuse enriches the visual context by concatenating the tokens generated by the vision encoders of selected MLLMs within a family. By merging the parameters of language models from these MLLMs, VisionFuse allows a single language model to align with various vision encoders, significantly reducing deployment overhead.
We conduct comprehensive evaluations across multiple multimodal benchmarks using various MLLM combinations, 
demonstrating substantial improvements 
in multimodal tasks. Notably, when integrating MiniGemini-8B and SLIME-8B, VisionFuse achieves an average performance increase of over 4\%.
\end{abstract}

\section{Introduction}
\begin{figure}[t]
  \centering
  \includegraphics[width=0.45\textwidth]{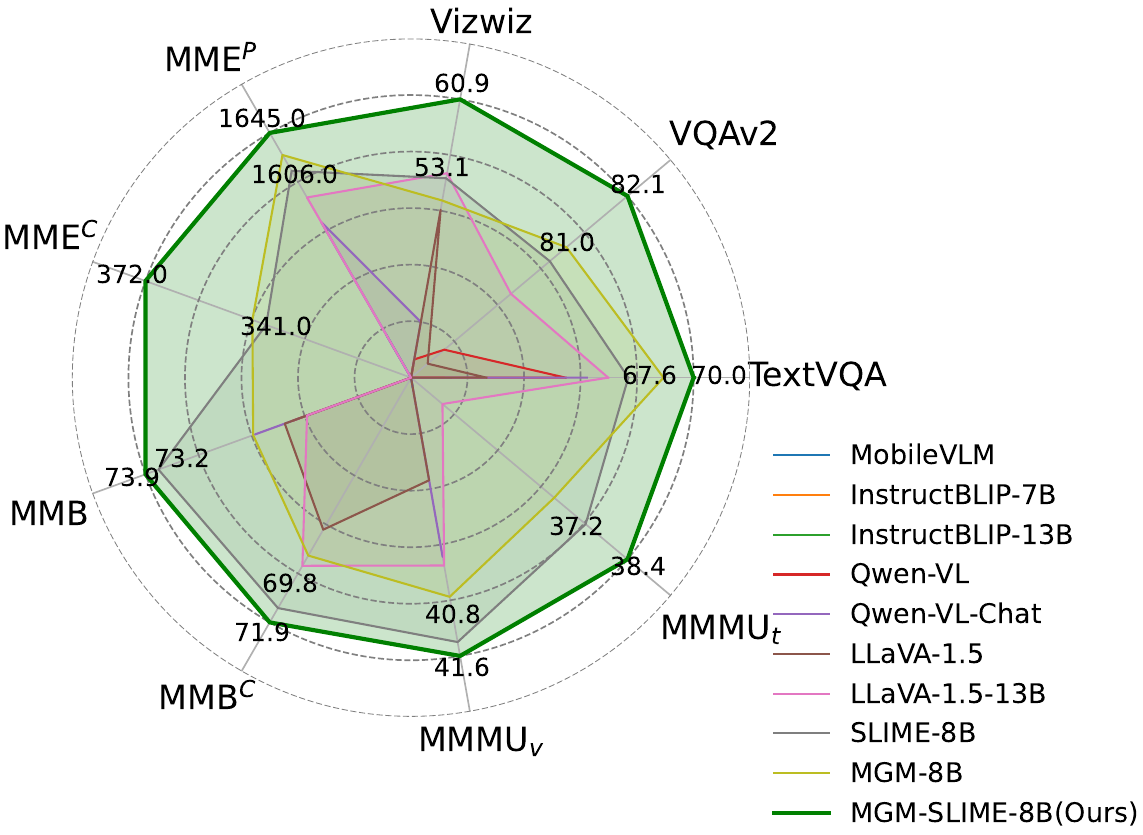}
  \includegraphics[width=0.45\textwidth]{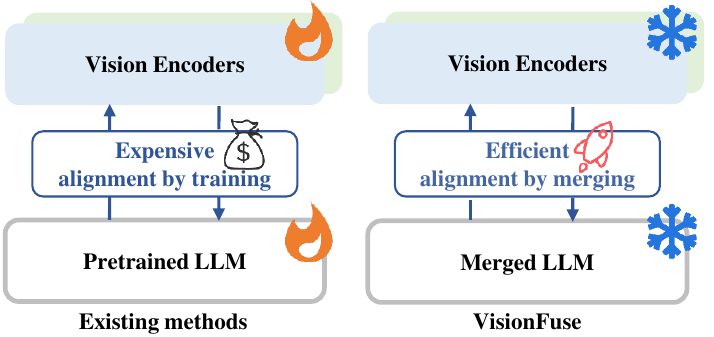}
   \caption{To enhance the perception capabilities of MLLMs, existing methods require substantial training costs to explore a vast design space and align each potential encoder with the language model. In contrast, our ~\methodname~ directly utilizes vision encoders from MLLMs within a family and aligns them with a single LLM by merging the parameters of LLMs, without incurring additional training overhead. For example, by integrating SLIME-8B and MGM-8B for free,~\methodname~ well exceeds the individual MLLMs and other leading methods.}
   \label{fig:radar}
\end{figure}

\begin{figure*}[t]
  \centering
    \includegraphics[width=\linewidth]{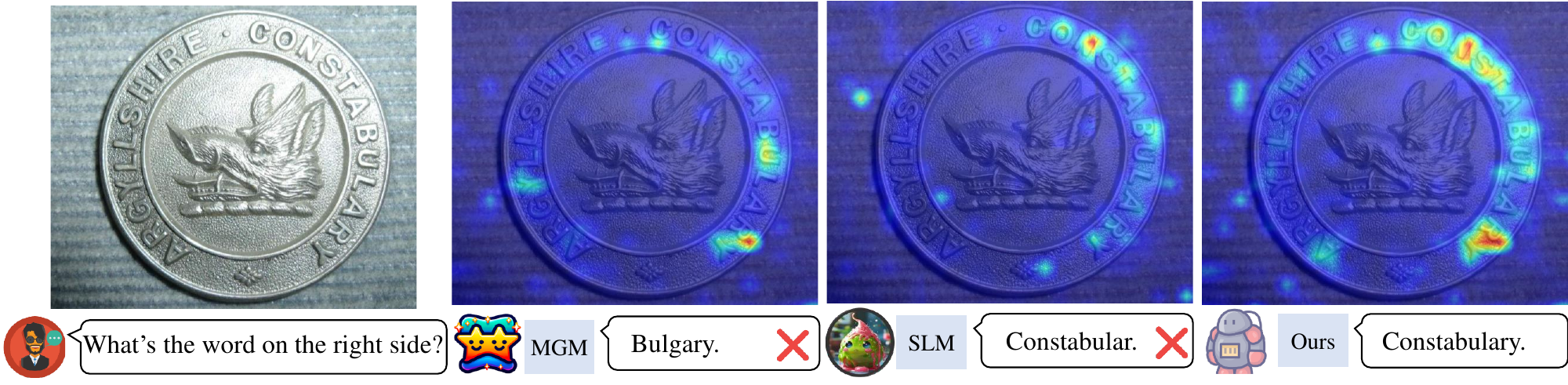}
\caption{\textbf{Different MLLMs exhibit varying visual perception capabilities.} We visualize the average cross-attention maps across all layers for two MLLMs - MGM and SLM, as well as for our method that integrates these two models, using an example to observe which areas the models focus on. It shows that our VisionFuse attention is more accurate, integrating the perceptual abilities of both MGM and SLM. 
Here, ``MGM" represents Mini-Gemini~\citep{li2024mini}, and ``SLM" represents SLIME~\citep{zhang2024beyond}.}
    \label{fig:motivation}
\end{figure*}

Multimodal LLMs (MLLMs) integrate vision encoders to Large Language Models (LLMs), allowing them to tackle multimodal tasks with emergent capabilities~\citep{liu2024visual,lin2024vila,ye2023mplug}. To handle complex and diverse multimodal tasks effectively, MLLMs require strong visual perception capabilities. A common approach to improving the visual perception of MLLMs is designing better vision encoders~\citep{cha2024honeybee,zhang2024beyond,shi2024eagle}. However, these methods typically involve switching vision encoders or altering image resolutions, both of which require 
re-align the vision encoders with the language model. 
Re-aligning individual encoders is highly time-consuming and computationally intensive, with training costs growing linearly with the number of configurations. For instance, aligning 10 vision encoders with a 7B language model using a data-efficient MLLM pipeline requires 3,840 NVIDIA A100 GPU hours, costing approximately \$20,000~\citep{yang2024law}. Evaluating all 1,023 combinations of these 10 encoders further increases development costs exponentially. Despite substantial effort, individual models may still face limitations in visual perception capabilities.


Recognizing these limitations, it becomes clear that different models tend to excel in distinct aspects of visual perception. For example, as illustrated in Figure~\ref{fig:motivation}, when asked the question, ``What's the word on the right side?", the models should focus on the entire word on the right side of the coin. MGM~\citep{li2024mini}, however, concentrates more on the lower right portion of the coin, with relatively less attention to the upper part. In contrast, SLIME~\citep{zhang2024beyond} directs more attention to the upper part, causing it to miss the last letter `y'. While MGM successfully recognizes the letter `y', its recognition of the preceding letters is suboptimal. Due to the perceptual limitations of both models, they provide incorrect answers in this text recognition task, yet demonstrate notable differences in their visual perception. This observation underscores the potential for mitigating the limitations of individual models by leveraging the complementary strengths of multiple models. By integrating the encoders in these MLLMs, our method captures a more complete target region, leading to accurate results.

To investigate this concept, ensemble learning has been proposed as an efficient means of leveraging the capabilities of different models without requiring additional training, by simply aggregating the outputs~\citep{jiang2023llm,wan2024knowledge,freitag2023epsilon}. However, deploying multiple full models and running inference on each one introduces inefficiencies in both memory consumption and computational resources. Since vision encoders are generally smaller and less resource-intensive than entire MLLMs, a viable alternative is to integrate multiple vision encoders with a single language model. For example, Eagle~\citep{shi2024eagle} enhances the visual perception of MLLMs by concatenating the encoded vision tokens from multiple encoders. While this approach reduces computational overhead compared to ensemble learning, it still encounters the challenge of aligning these encoders with a single language model, which often necessitates multimodal instruction tuning and imposes considerable computational costs~\citep{yang2024law}.

\begin{figure*}[t]
  \centering
    \includegraphics[width=1.0\linewidth]{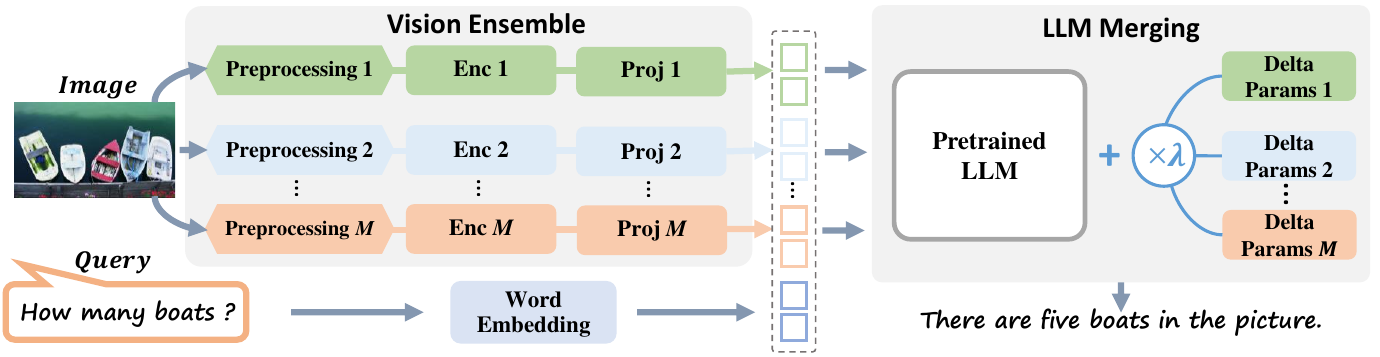}
    \caption{Overview of \methodname. \methodname~merges the language model parameters from $M$ different pretrained MLLMs within a family to align a single language model with multiple vision encoders. The merged LLM is obtained by the weighted linear interpolation of the pretrained LLM and $M$ delta parameters, which are the changes in parameters of LLMs during fine-tuning. The input image is processed through distinct preprocessing pipelines, consistent with those in MLLMs, as well as vision encoders and projectors, to extract richer visual features. These features are then concatenated with text tokens and fed into the merged language model.}
    \label{fig:method}
\end{figure*}

In this paper, we introduce \methodname, a novel framework designed to efficiently enhance the visual perception capabilities of MLLMs, as illustrated in Figure~\ref{fig:method}. We define the MLLM family as a group of models that share the same pre-trained language model. For example, both MiniGemini-8B~\citep{li2024mini} and SLIME-8B~\citep{zhang2024beyond} are trained using LLaMA-3-8B-Instruct~\citep{llama3modelcard}, and therefore belong to the same MLLM family. We first conduct a statistical analysis of cross-attention to highlight variations in visual perception across different MLLMs, and then propose enhancing visual perception by integrating these models. Notably, we observe that vision encoders within an MLLM family exhibit similar feature distributions, making their tokens more compatible for combination. 
Furthermore, we find that merging language models within an MLLM family effectively aligns a single LLM with different vision encoders, so prior to deployment, we merge language model parameters from various MLLMs to achieve this alignment.

During inference, we apply preprocessing pipelines to the input visual data consistent with those used by individual MLLMs within the family, then feed the processed data into the vision encoders and projectors from each MLLM to extract visual tokens.
These tokens are then concatenated to enrich contextual information. \methodname~harnesses the visual perception capabilities of multiple MLLMs, enhancing multimodal task performance with minimal additional inference overhead from the vision encoders.

We apply~\methodname~to different MLLM families, based on the pretrained language models including Vicuna-v1.5~\citep{chiang2023vicuna} and LLaMA-3-8B-instruct~\citep{llama3modelcard}. Compared to the individual model baselines, our approach demonstrates significant improvements across multiple multimodal datasets. Furthermore, we visualize the cross-attention to intuitively show that the model after integrating focuses on more relevant positions than the individual models.

In summary, our contributions are as follows:

\begin{itemize}[leftmargin=*]
    \item \textbf{New insights in different MLLMs}: We identify three key phenomena in various pretrained MLLMs: (1) different models focus on distinct image regions for the same input, (2) vision encoders within an MLLM family exhibit more consistent features, and (3) merging language model parameters is critical for aligning the language model with different vision encoders.
    
    \item \textbf{A training-free method to enhance the perception capabilities of MLLM}: We propose \methodname, a simple yet effective method for integrating MLLMs, based on three key observations. By merging language model parameters to align with different vision encoders and leveraging vision encoders from multiple MLLMs within a family,~\methodname~enhances visual perception with minimal deployment overhead. Extensive experiments demonstrate that~\methodname~effectively enhances performance on multimodal tasks. Specifically, in the integration of Mini-Gemini-8B and SLIME-8B,~\methodname~achieves a significant average improvement of over 4\% across multiple multimodal benchmarks without additional training.
\end{itemize}

\section{Related Work}

\textbf{Enhancing the visual perception of MLLMs.} Existing multimodal large language models (MLLMs) primarily enhance their visual perception by incorporating high-resolution inputs~\citep{li2024mini}, employing optimized preprocessing methods to capture richer visual features~\citep{liu2024llavanext,zhang2024beyond}, and designing more effective vision modules~\citep{zhang2024beyond,cha2024honeybee,ge2024convllava}. Specifically, LLaVA-Next~\citep{liu2024llavanext} segments input images into local patches and uses high-quality data to train the MLLM. Mini-Gemini~\citep{li2024mini} uses CLIP~\citep{radford2021clip} tokens as low-resolution queries to cross-attend to another high-resolution vision encoder within co-located local windows. Honeybee~\citep{cha2024honeybee} introduces a locality-enhanced projector that balances token management flexibility with local visual context preservation, improving both efficiency and performance in spatial understanding tasks. ConvLLaVA~\citep{ge2024convllava} leverages a hierarchical ConvNeXt backbone to compress high-resolution images into fewer visual tokens, enhancing efficiency while maintaining spatial understanding across diverse image resolutions. SliME~\citep{zhang2024beyond} refines visual adapters by employing a mixture of experts for global features and compressing local image tokens with query embeddings, improving both efficiency and performance in high-resolution tasks. Eagle~\citep{shi2024eagle} explores the design space of multimodal LLMs by integrating multiple vision encoders with different architectures and pretraining tasks, enhancing multimodal performance through efficient fusion strategies like direct token concatenation. However, these approaches require extensive fine-tuning to align the language model with the vision modules, leading to significant computational costs. Similar to existing methods~\citep{shi2024eagle}, our approach also concatenates features from multiple visual encoders to enhance the visual perception capabilities of the model. In contrast, our method efficiently improves the visual perception abilities of multimodal large language models without requiring extensive additional training.

\begin{figure*}[t]
    \centering
    \begin{subfigure}[b]{0.330\textwidth}
        \centering
        \includegraphics[width=\textwidth]{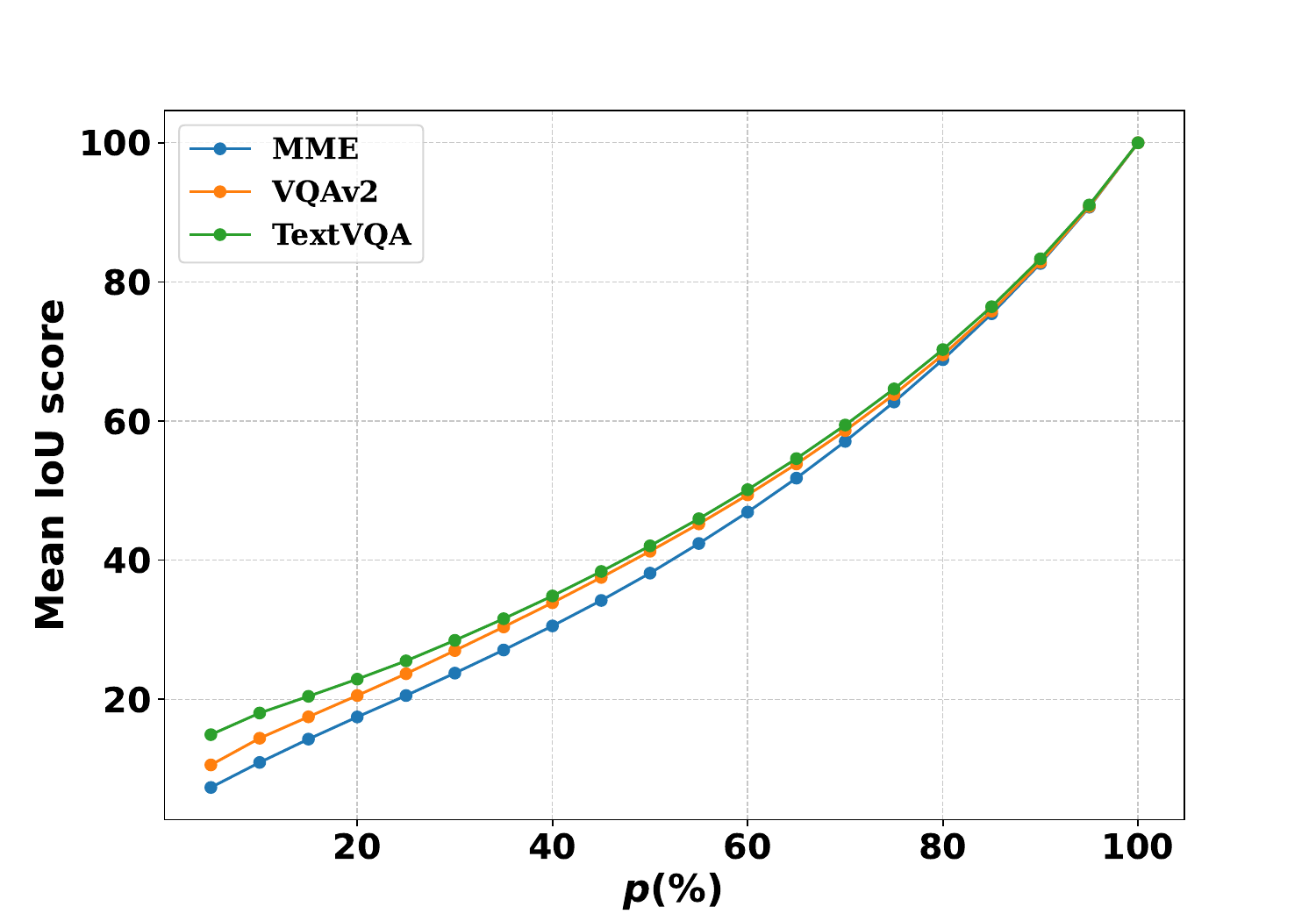}
        \caption{IoU of top $p\%$ vision tokens with highest cross attention in different MLLMs.}
        \label{fig:iou_crossattn}
    \end{subfigure}
    \hfill
    \begin{subfigure}[b]{0.327\textwidth}
        \centering
        \includegraphics[width=\textwidth]{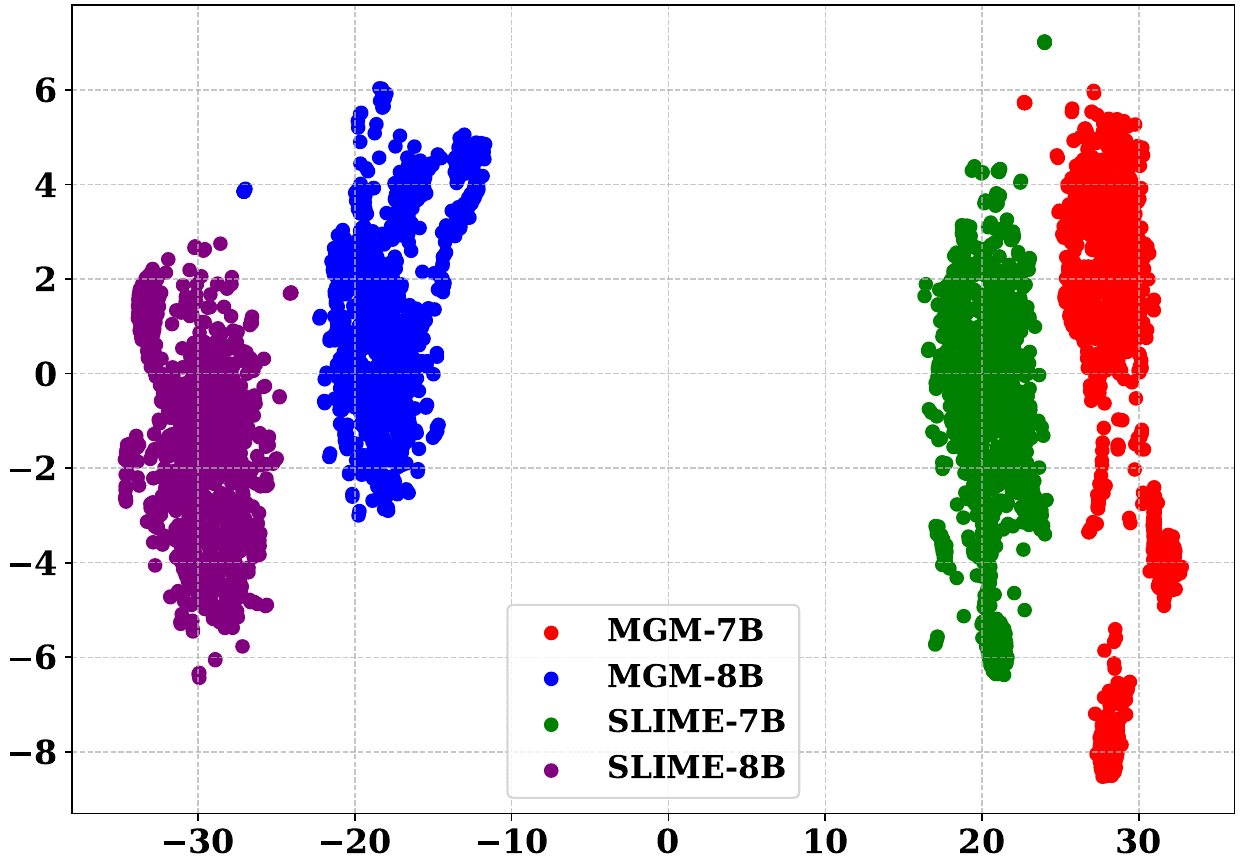}
        \vspace{-0.3cm}
        \caption{Visualizations of visual feature distributions of four multimodal models.}
        \label{fig:tsne}
    \end{subfigure}
    \hfill
    \begin{subfigure}[b]{0.327\textwidth}
        \centering
        \includegraphics[width=\textwidth]{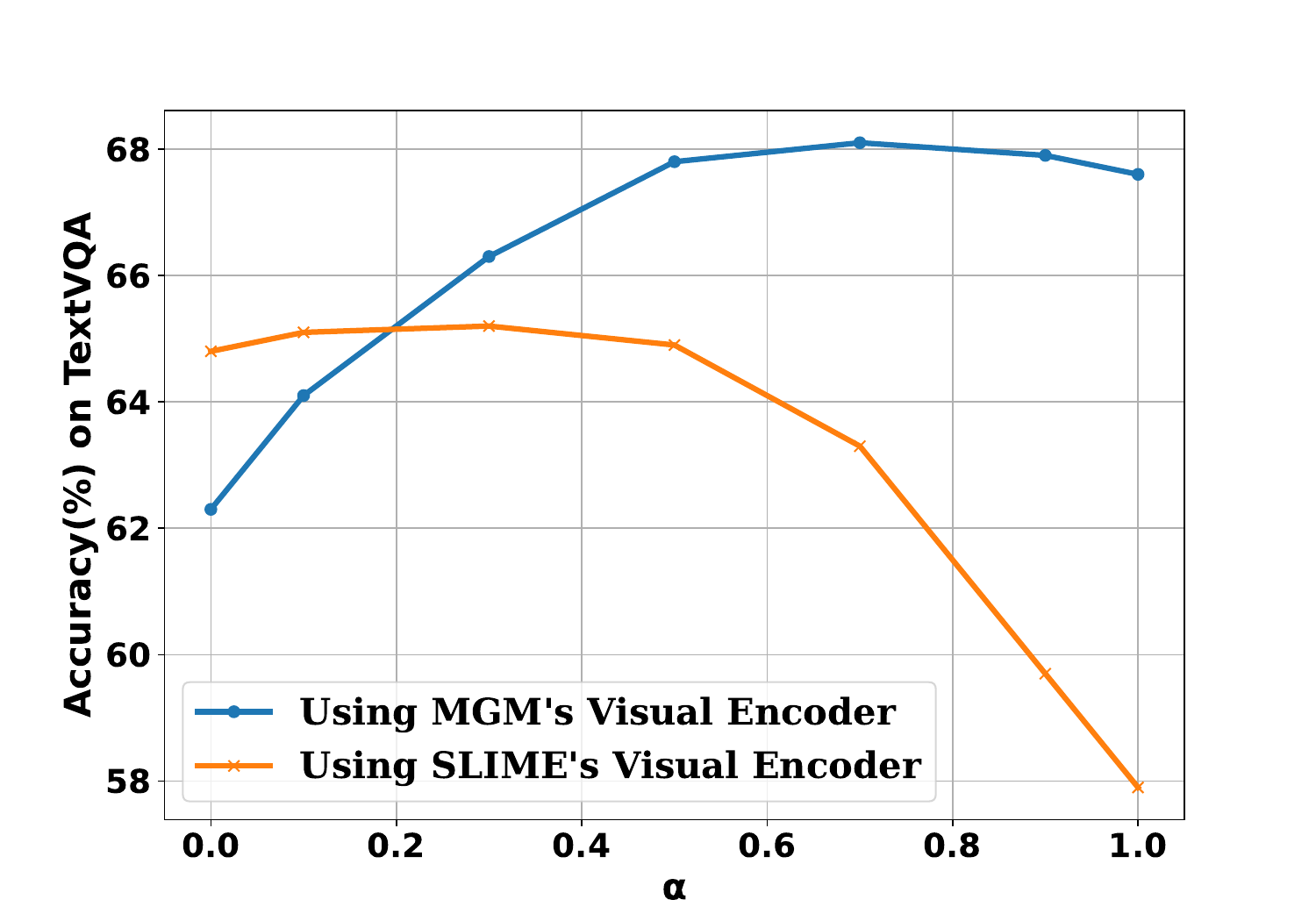}
        \vspace{-0.4cm}
        \caption{Accuracy on TextVQA when interpolating MGM and SLIME with different $\alpha$.}
        \label{fig:tv_motivation}
    \end{subfigure}
    \caption{Summary of our exploration and observations: (a) demonstrates that different MLLMs focus on distinct image regions for the same visual and textual inputs; (b) reveals that vision encoders within an MLLM family exhibit more similar feature distributions; and (c) highlights the importance of merging language model parameters to align the language model with different vision encoders.}
    \label{fig:three_images}
\end{figure*}

\noindent\textbf{Model Merging.} Model merging seeks to consolidate multiple parameter sets into a single model without requiring retraining, providing a more memory-efficient and cost-effective alternative to model ensembling by eliminating the need to store multiple checkpoints. Existing methods for merging parameters of LLMs generally fall into two categories: merging with coefficients and parameter sparsification. Task Arithmetic~\citep{DBLP:conf/iclr/IlharcoRWSHF23} introduces task vectors to efficiently edit pre-trained models by performing arithmetic operations on weight differences, enhancing task performance without retraining. Regmean~\citep{jin2022dataless} computes a closed-form solution for the merging coefficients by minimizing the difference in activation values before and after merging. SLERP~\citep{shoemake1985animating} determines the merging coefficients for parameter addition based on the angle between the model parameters. Methods like Ties-merging~\citep{yadav2023resolving} and DARE~\citep{yu2024language} observe that the changes of in parameters of LLMs during fine-tuning contain a significant amount of redundancy, and propose reducing conflicts between different delta parameters through pruning. However, these model merging methods are primarily designed for language or vision models to preserve the capabilities of individual models across different downstream tasks. Our method aligns language models with different vision encoders through model merging, which is significantly different from the motivations of existing model merging approaches.

\section{Empircal Insights}
\label{pilot_study}
In this section, we provide comprehensive visualizations and discussions from the following perspectives to introduce three novel insights. The experiments are conducted using the following models: SLIME-7B~\citep{zhang2024beyond} and MGM-7B~\citep{li2024mini}, both based on Vicuna-v1.5~\citep{chiang2023vicuna}, as well as SLIME-8B~\citep{zhang2024beyond} and MGM-8B~\citep{li2024mini}, which are based on LLaMA-3-8B-Instruct~\citep{llama3modelcard}. We evaluate these models on the following datasets: TextVQA~\citep{singh2019towards}, MME~\citep{fu2024mmebenchmark}, and VQAv2~\citep{goyal2017making}.

\noindent\textbf{Observation 1: Different MLLMs attend to different regions for the same query and visual input.} To investigate the regions of focus for different MLLMs when given the same visual input and query, we analyze SLIME-7B and MGM-7B using samples from TextVQA, MME, and VQAv2. We calculate the average attention across all layers and the average Intersection over Union (IoU) of the top $p\%$ regions (tokens) with the highest cross-attention. As shown in Figure~\ref{fig:iou_crossattn}, even with the same visual input and query, different models tend to focus on distinct regions, indicating notable differences in visual perception capabilities across multimodal models. For instance, in the MME dataset, the IoU of the top 5\% of tokens with the highest cross-attention score is less than 10\%. As demonstrated in Figure~\ref{fig:motivation}, there are significant differences in the cross-attention between MGM and SLIME when presented with the same image and text query. \textit{Thus, we propose leveraging these differences to provide the language model with richer visual information.}

\noindent\textbf{Observation 2: Visual feature distributions of encoders within an MLLM family exhibit a closer alignment.}
MLLMs align visual and textual features during multimodal instruction fine-tuning, allowing visual features to serve as additional contextual information for textual input. Intuitively, vision encoders within an MLLM family should have more closely aligned visual features, as they are aligned with the same pretrained language model. To validate this hypothesis, we randomly select 100 samples from the TextVQA~\citep{singh2019towards} dataset and visualize the distribution of all visual tokens from different vision encoders using t-SNE~\citep{van2008tsne}, focusing on SLIME-7B, MGM-7B~\citep{li2024mini}, SLIME-8B, and MGM-8B. As illustrated in Figure~\ref{fig:tsne}, feature distributions of vision encoders within the same MLLM family are relatively similar, whereas those trained on different language models display more distinct distributions. This indicates that \textit{vision encoders within an MLLM family tend to generate more closely aligned visual features, making them more compatible for integration within a shared context}.

\noindent\textbf{Observation 3: Merging language model parameters aligns the language model with different vision encoders.} Previous research on model merging~\citep{ilharco2022editing,yadav2023resolving} has primarily focused on combining delta parameters to enable a single LLM to inherit capabilities from multiple fine-tuned models. This motivate us to explore whether merging the delta parameters of the language model, relative to the pretrained language model in MLLMs, similarly facilitates alignment between the language model and different vision encoders. Using SLIME-8B and MGM-8B as examples, we calculate their delta parameters relative to LLaMA-3-8B-Instruct~\citep{li2023llama} and merge them via linear interpolation, formalized as follows:
\begin{align}
    \bm{\Theta_\text{interpolate}} &= \alpha \cdot (\bm{\Theta_\text{MGM}} - \bm{\Theta_\text{LLaMA3}}) \nonumber \\
    &\quad + (1 - \alpha) \cdot (\bm{\Theta_\text{SLIME}} - \bm{\Theta_\text{LLaMA3}}) \nonumber \\
    &\quad + \bm{\Theta_\text{LLaMA3}},
\label{eq:interpolation}
\end{align}
where $\bm{\Theta_\text{interpolate}}$ represents the parameters after interpolation, $\bm{\Theta_\text{MGM}}$, $\bm{\Theta_\text{SLIME}}$, and $\bm{\Theta_\text{LLaMA3}}$ represent the parameters of MGM-8B, SLIME-8B, and LLaMA-3-8B-Instruct, respectively. We evaluate the performance of the merged delta parameters on TextVQA. As shown in Figure~\ref{fig:tv_motivation}, when $\alpha$ is set to 0 or 1, the model corresponds to SLIME-8B or MGM-8B, respectively. However, when $\alpha$ is closer to 0.5, the model achieves optimal performance by effectively utilizing both vision encoders. This demonstrates that the delta parameters between an MLLM and its base model are key to enabling a single language model to align with different vision encoders. \textit{By merging delta parameters from different MLLMs within a family, we can align a single language model with multiple vision encoders.}
\section{Methodology}
\label{sec:method}

Inspired by our observations, we propose \methodname, a simple yet effective approach for efficiently integrating different MLLMs to enhance visual perception. As illustrated in Figure~\ref{fig:method} and Algorithm~\ref{alg:method}, ~\methodname~first merges the language models and then utilizes various vision encoders to extract richer features for the input image, which are subsequently fed into the merged LLM.

\subsection{Preliminaries}

\textbf{Notation}: Let $f_{\bm{\Theta}}(x)$ represent the language model of an MLLM, where $\bm{\Theta}$ denotes its parameters. The input $x$ consists of a sequence of tokens, including vision tokens $V$ from the modality-specific encoder and text tokens $T$ from word embeddings. The parameters of the pre-trained language model are denoted as $\bm{\Theta_{\text{pre}}}$.

\noindent\textbf{Delta parameters.} Delta parameters represent the changes in model parameters during fine-tuning~\citep{liu2024bitdelta}. In MLLMs, the delta parameters for the language model during multimodal fine-tuning are expressed as $\bm{\Theta - \Theta_{\text{pre}}}$.

\noindent\textbf{Main components of the MLLM.} Existing MLLMs~\citep{yin2023survey} typically consist of a modality preprocessing module that processes input data (e.g., slicing images into local patches), a modality-specific encoder that converts the data into features, a modality-specific projector that maps the encoded visual features into the text space, and an LLM that performs cross-modal reasoning by integrating these tokens with text-based input. For a given set of $M$ MLLMs, the language model parameters of the $i$-th model are denoted as $\bm{\Theta_\text{i}}$, its preprocessing pipeline as $\text{Preprocessing}_i$, its vision encoder as $\text{Enc}_i$, and its vision projector as $\text{Proj}_i$.

\begin{algorithm}[t]
\caption{Procedure of Inference for VisionFuse}
\label{alg:method}
\begin{algorithmic}
\Require $M$ MLLMs built upon the same pretrained language model with parameters $\bm{\Theta_{\text{pre}}}$, with the $i$-th model having language model's parameter $\bm{\Theta_\text{i}}$, vision encoder $\text{Enc}_i$ and preprocessing pipeline $\text{Preprocessing}_i$, input image $x$, text input $T$.
\Ensure Prediction $\hat{y}$
\State Generate merged parameters $\bm{\Theta_{\text{merged}}}$ using Eq.~(\ref{eq:tv}). \Comment{Merge language model parameters}
\For{the $i$-th vision encoder from $1$ to $M$}  \Comment{Extract visual features from different encoders}
    \State Extract visual features $V_i$ using Eq.~(\ref{eq:encoder}).
\EndFor
\State Generate $V_F$ using Eq.~(\ref{eq:concatenation}).
\State \Return Prediction $\hat{y}$ using Eq.~(\ref{eq:prediction}).
\end{algorithmic}
\end{algorithm}

\label{headings}

\subsection{Ensemble of Different Vision Encoders}
To leverage the visual perception capabilities of different MLLMs, an intuitive approach is to combine multiple MLLMs into an ensemble and aggregate their predictions. However, due to the large number of parameters in the language models, directly using an ensemble of the entire MLLMs would result in significant computational overhead.
Therefore, we explore whether it is feasible to integrate only the vision encoders, which have relatively fewer parameters, and feed the output tokens into the language model for inference.

In pre-trained MLLMs, the alignment between textual and visual inputs in the feature space allows them to be treated as a unified sequence for input into the language model. As discussed in \textbf{Observation 2}, visual features from MLLMs trained on the same language model are more closely aligned in the feature space. Therefore, we directly aggregate the outputs of different vision encoders within an MLLM family, treating them as distinct visual context information, which enables the language model to obtain richer visual perception. Due to the varying lengths of vision tokens from different MLLMs, we propose directly concatenating the vision tokens from these MLLMs. The process of combining multiple vision encoders into an ensemble can be formalized as follows.:
\begin{equation}
V_i = \text{Proj}_i(\text{Enc}_i(\text{Preprocessing}_i(x))), \label{eq:encoder}
\end{equation}
\begin{equation}
V_F = [V_1; V_2; \dots; V_n],\label{eq:concatenation}
\end{equation}
where $[\cdot;\cdot]$ denotes concatenation, and $V_F$ represents the integrated visual features.

\subsection{Merging LLMs from a Family of MLLMs}
A single language model cannot directly align with multiple encoders from different MLLMs, as they have not undergone alignment training. Retraining for such alignment would result in substantial computational costs. As discussed in \textbf{Observation 3}, merging language model parameters within an MLLM family helps align a language model with different vision encoders. Therefore, following the approach in~\citep{ilharco2022editing}, we merge these delta parameters to create a single language model capable of interpreting visual tokens from multiple vision encoders. 
The merging process
can be formalized as follows:

\vspace{-0.4cm}
\begin{equation}
\bm{\Theta_{\text{merged}}} = \bm{\Theta_{\text{pre}}} + \lambda \cdot \sum_{i=1}^{M}{(\bm{\Theta_\text{i}} - \bm{\Theta_{\text{pre}}})},
\label{eq:tv}
\end{equation}

\noindent
where $\bm{\Theta_{\text{pre}}}$ represents the parameters of the shared base model, and $\lambda$ is a merging coefficient. The prediction $\hat{y}$ can then be generated as follows:

\vspace{-0.0cm} 
\begin{equation}
\hat{y} = f(V_F, T; \bm{\Theta_{\text{merged}}}).
\label{eq:prediction}
\end{equation}

\noindent\textbf{Complexity analysis.} 
\textit{For training complexity, our \methodname~is significantly more cost-effective than existing methods.}
Existing MLLMs require extensive training cost to achieve alignment between the language model and vision encoders \citep{yang2024law}. 
Instead, \methodname~enhances the perception capabilities by concatenating tokens from different vision encoders and merging parameters of the language models from different MLLMs, with no additional training cost.

\textit{For inference complexity, our \methodname~ requires some additional cost to process the longer visual tokens.}
Since the computational cost of the language model of MLLMs is significantly higher than that of the vision encoder, we mainly analyze the computational cost of the language model for different methods.
The Floating Point Operations (FLOPs) in language model for $i$-th MLLM are $O\left(\left( L_{\text{Enc}}^i + L_t\right)^2\right)$, where $L_{\text{Enc}}^{i}$ represents the length of generated tokens from $i$-th MLLM's vision encoders, $L_t$ represents the length of text tokens.
Since our \methodname~concatenates the tokens from all $M$ visual modules of MLLMs, it requires processing more tokens compared to a single MLLM. Specifically, the FLOPs of \methodname~are $O\left(\left(\sum_{i=1}^{M} L_{\text{Enc}}^i + L_t\right)^2\right)$.
To reduce the inference cost, one can use the token pruning method to remove the redundant visual tokens~\citep{chen2024fastv}. 
We analyze our method under different levels of sparsity and the results demonstrate that \textit{our method is able to achieve better performance with lower FLOPs compared to a single model}. Detailed discussions of efficiency are included in the appendix.
\section{Experiments}
\label{others}

\textbf{Implementation details.} We conduct experiments across various MLLM combinations, including (1) SLIME-7B~\citep{zhang2024beyond} and MGM-7B~\citep{li2024mini} based on Vicuna-v1.5~\citep{chiang2023vicuna}, and (2) SliME-8B~\citep{zhang2024beyond} and MGM-8B~\citep{li2024mini} based on Llama-3-8B-Instruct~\citep{llama3modelcard}. We evaluate the performance of our approach on multiple multimodal datasets, including VQA\(^T\) (TextVQA)~\citep{singh2019towards}, MMB (MMBench)~\citep{liu2023mmbench}, MMB\(^C\)(MMBench-Chinese)~\citep{liu2023mmbench}, MME~\citep{fu2024mmebenchmark}, MMMU~\citep{yue2024mmmu}, VQAv2~\citep{goyal2017vqav2} and Vizwiz~\citep{gurari2018vizwiz}.

\noindent\textbf{Compared methods.} We compare our method with the baselines and existing leading MLLMs, including MobileVLM~\citep{chu2023mobilevlm}, Qwen-VL~\cite{bai2023qwen}, Qwen-VL-Chat~\cite{bai2023qwen}, IDEFICS~\citep{laurencon2023introducing}, LLaMA-VID~\citep{li2023llama}, LLaVA-1.5~\citep{liu2024improved}, VILA~\citep{lin2024vila}, Shika~\citep{chen2023shikra} and InstructBLIP~\citep{dai2023instructblip}.

\begin{table*}[ht]
\caption{Comparison with leading methods on multimodal benchmarks. Results of~\methodname~are marked in gray. VQA\(^T\): TextVQA; MMB: MMBench; MMB\(^C\): MMBench-Chinese; MMMU\(_{v,t}\): validation and test set of MMMU; MME\(^{P,C}\): Perception and Cognition in MME. Res. indicates the input resolution. Percentages indicate the rate of improvement compared to the best performance of the baselines.
}
\label{tab:main_results}
\centering
\small
\renewcommand{\tabcolsep}{2.0pt}
\begin{tabular}{l|l|c|c|c|c|c|c|c|c|c|c}
\toprule
\textbf{Method} & \textbf{LLM} & \textbf{Res.} & \textbf{VQA\(^T\)} & \textbf{VQAv2} & \textbf{Vizwiz} & \textbf{MME\(^P\)} & \textbf{MME\(^C\)} & \textbf{MMB} & \textbf{MMB\(^C\)} & \textbf{MMMU$_v$} & \textbf{MMMU$_t$} \\ 
\midrule
MobileVLM & MLLaMA 2.7B & 336 & 47.5 & - & - & 1289 & - & 59.6 & - & 26.2 & - \\ 
InstructBLIP & Vicuna-7B & 224 & 50.1 & - & 34.5 & - & - & - & 36.0 & - & - \\ 
InstructBLIP & Vicuna-13B & 336 & 50.7 & - & 33.4 & 1213 & - & - & - & 25.6 & - \\ 
Qwen-VL & Qwen-7B & 336 & 59.8 & 78.8 & 35.2 & - & - & 66.0 & - & - & - \\ 
Qwen-VL-Chat & Qwen-7B & 448 & 61.5 & 78.2 & 38.9 & 1488 & - & 68.0 & - & 35.9 & 32.2 \\ 
Shikra & Vicuna-13B & 336 & 52.3 & - & - & - & - & 59.2 & - & - & - \\ 
IDEFICS-80B & LLaMA-65B & 224 & 30.9 & - & - & - & - & 54.5 & - & - & - \\ 
LLaMA-VID & Vicuna-7B & 336 & - & 79.3 & 54.2 & 1521 & - & 65.1 & - & - & - \\ 
LLaMA-VID & Vicuna-13B & 336 & - & 80.0 & 54.3 & 1521 & - & 66.6 & - & - & - \\ 
LLaVA-1.5 & Vicuna-7B & 336 & 53.5 & 78.5 & 50.0 & - & - & 66.4 & 58.3 & 31.5 & - \\ 
LLaVA-1.5 & Vicuna-13B & 336 & 63.2 & 80.0 & 53.6 & 1531 & 295 & 65.2 & 63.6 & 36.4 & 33.1 \\ 
LLaVA-HD & Vicuna-13B & 336 & 62.5 & 81.8 & 57.5 & 1500 & - & 68.8 & 61.9 & - & - \\ 
MGM-8x7B & Mixtral-8x7B & 336 & 69.2 & - & - & 1639 & 379 & 75.6 & - & 41.8 & 37.1 \\ 
\midrule
MGM-7B & Vicuna-7B & 336 & 65.2 & 80.4 & 52.1 & 1523 & 316 & 68.7 & 57.8 & 36.1 & 32.8 \\ 
SliME-7B & Vicuna-7B & 336 & 64.4 & 80.3 & 53.7 & 1544 & 383 & 68.4 & 61.3 & 37.2 & 33.4 \\ 
\rowcolor{gray!20}
 &  &  & \textbf{66.9} & \textbf{80.7} & \textbf{54.4} & \textbf{1563} & \textbf{394} & \textbf{69.6} & \textbf{62.5} & \textbf{37.8} & \textbf{33.6} \\ 
\rowcolor{gray!20}
\multirow{-2}{*}{MGM-SliME-7B} & \multirow{-2}{*}{Vicuna-7B} & \multirow{-2}{*}{336} & \textcolor{blue}{+2.6\%} & \textcolor{blue}{+0.4\%} & \textcolor{blue}{+1.3\%} & \textcolor{blue}{+1.2\%} & \textcolor{blue}{+2.9\%} & \textcolor{blue}{+1.3\%} & \textcolor{blue}{+2.0\%} & \textcolor{blue}{+1.6\%} & \textcolor{blue}{+0.6\%} \\ 

\midrule
MGM-8B & LLaMA-3-8B-Instruct & 336 & 67.6 & 81.0 & 50.9 & 1606 & 341 & 68.1 & 62.1 & 38.2 & 36.3 \\ 
SliME-8B & LLaMA-3-8B-Instruct & 336 & 64.8 & 80.7 & 53.1 & 1578 & 337 & 73.2 & 69.8 & 40.8 & 37.2 \\ 
\rowcolor{gray!20}
 &  &  & \textbf{70.0} & \textbf{82.1} & \textbf{60.9} & \textbf{1645} & \textbf{372} & \textbf{73.9} & \textbf{71.9} & \textbf{41.6} & \textbf{38.4} \\ 
 \rowcolor{gray!20}
\multirow{-2}{*}{MGM-SliME-8B} & \multirow{-2}{*}{LLaMA-3-8B-Instruct} & \multirow{-2}{*}{336} & \textcolor{blue}{+3.6\%} & \textcolor{blue}{+1.4\%} & \textcolor{blue}{+14.7\%} & \textcolor{blue}{+2.4\%} & \textcolor{blue}{+9.1\%} & \textcolor{blue}{+1.0\%} & \textcolor{blue}{+3.0\%} & \textcolor{blue}{+2.0\%} & \textcolor{blue}{+3.2\%} \\

\midrule
MGM-8B-HD & LLaMA-3-8B-Instruct & 672 & 71.6 & 81.5 & 54.4 & 1532 & 357 & 70.6 & 65.2 & 37.0 & 36.5 \\ 
SliME-8B & LLaMA-3-8B-Instruct & 336 & 64.8 & 80.7 & 53.1 & 1578 & 337 & 73.2 & 69.8 & 40.8 & 37.2 \\ 
\rowcolor{gray!20}
 &  &  & \textbf{72.7} & \textbf{82.2} & \textbf{55.2} & \textbf{1600} & \textbf{364} & \textbf{75.1} & \textbf{70.8} & \textbf{41.8} & \textbf{37.4} \\ 
\rowcolor{gray!20}
\multirow{-2}{*}{MGM-HD-SliME-8B} & \multirow{-2}{*}{LLaMA-3-8B-Instruct} & \multirow{-2}{*}{672} & \textcolor{blue}{+1.5\%} & \textcolor{blue}{+0.9\%} & \textcolor{blue}{+1.5\%} & \textcolor{blue}{+1.4\%} & \textcolor{blue}{+2.0\%} & \textcolor{blue}{+2.5\%} & \textcolor{blue}{+1.4\%} & \textcolor{blue}{+2.5\%} & \textcolor{blue}{+0.5\%} \\

\bottomrule
\end{tabular}
\end{table*}

\noindent\textbf{Main results.} We evaluate our method across several multimodal datasets and compare it against the leading MLLMs, as detailed in Table~\ref{tab:main_results}. Without any additional training, our approach significantly enhances performance over individual models by simply integrating vision encoders from the same MLLM family in both the the combinations based on Vicuna-7B and LLaMA-3-8B-Instruct. Notably, in the integration of MGM-8B and SLIME-8B, ~\methodname~incurs only a 3.4\% increase in parameters due to the additional encoders employed, achieving a 4\% relative improvement compared to the optimal individual model. Furthermore, the performance is on par with that of the MGM-8x7B model, which contains over six times more parameters, highlighting the parameter efficiency of~\methodname. The results of integrating more MLLMs are included in the appendix.

\noindent\textbf{Effectiveness on different resolutions inputs.} To further assess the effectiveness of our method with varying image resolutions, we combine the high-resolution vision encoder from Mini-Gemini-HD-8B with the low-resolution vision encoder from SLIME-8B. As demonstrated in Table~\ref{tab:main_results}, even the low-resolution vision encoder can mitigate some of the limitations of the high-resolution encoder by providing richer visual information to the language model. The performance after integration is higher than that of either individual model. This suggests that even models utilizing high-resolution inputs may overlook critical regions, whereas the low-resolution vision encoder can help address these gaps in visual perception.

\begin{table*}[ht]
\caption{Ablation study on each component. We conduct ablation experiments on the merging of MGM-8B and SLIME-8B. In this context, `V.E. of SLIME' refers to using the vision encoder from SLIME-8B, and `V.E. of MGM' refers to using the vision encoder from MGM-8B. `SLIME's Param' indicates the use of SLIME-8B as the language model, and `MGM's Param' indicates the use of MGM-8B as the language model. When both `MGM's Param' and `SLIME's Param' are selected, we merge the parameters of MGM-8B and SLIME-8B.}
\label{tab:ablation}
\centering
\small
\renewcommand{\tabcolsep}{3.0pt}
\begin{tabular}{cccc|ccccc}
\toprule
\textbf{SLIME's Param} & \textbf{MGM's Param} & \textbf{V.E. of SLIME} & \textbf{V.E. of MGM} & \textbf{TextVQA} & \textbf{MME\(^P\)} & \textbf{MME\(^E\)} & \textbf{GQA} & \textbf{POPE} \\
\midrule
\checkmark & \checkmark & \checkmark & \checkmark & \textbf{70.0} & \textbf{1645} & \textbf{372} & \textbf{64.3} & \textbf{85.8} \\
\midrule
  & \checkmark & & \checkmark & 67.6 & 1606 & 341 & 63.2 & 85.6 \\
 \checkmark &  & \checkmark & & 64.8 & 1578 & 337 & 63.9 & 84.9 \\
 & \checkmark & \checkmark & \checkmark & 66.4 & 1485  & 294 & 62.3 & 84.2 \\
 \checkmark & & \checkmark & \checkmark & 66.9 & 1590 & 355 & 62.5 & 84.6 \\
 \checkmark & \checkmark & \checkmark &  & 65.9  & 1606  & 358 & 63.1 & 83.1 \\
 \checkmark & \checkmark &  & \checkmark & 68.1 &  1628 & 348 & 63.5 & 82.9 \\
\bottomrule
\end{tabular}
\end{table*}

\noindent\textbf{Effectiveness of each component.} To assess the effectiveness of each component of our method, we perform ablation studies on several multimodal datasets, as shown in Table~\ref{tab:ablation}. The results demonstrate that without merging the delta parameters, the language model fails to align with the vision encoders from different MLLMs, leading to a decline in performance. On the other hand, merging the delta parameters without incorporating features from multiple vision encoders results in moderate improvements in multimodal tasks. However, these gains are limited by the absence of richer visual information. Therefore, the substantial performance enhancement is largely attributed to the inclusion of richer visual information.

\begin{figure*}[t]
    \centering
    \begin{minipage}{0.31\textwidth}
        \centering
        \includegraphics[width=0.9\linewidth]{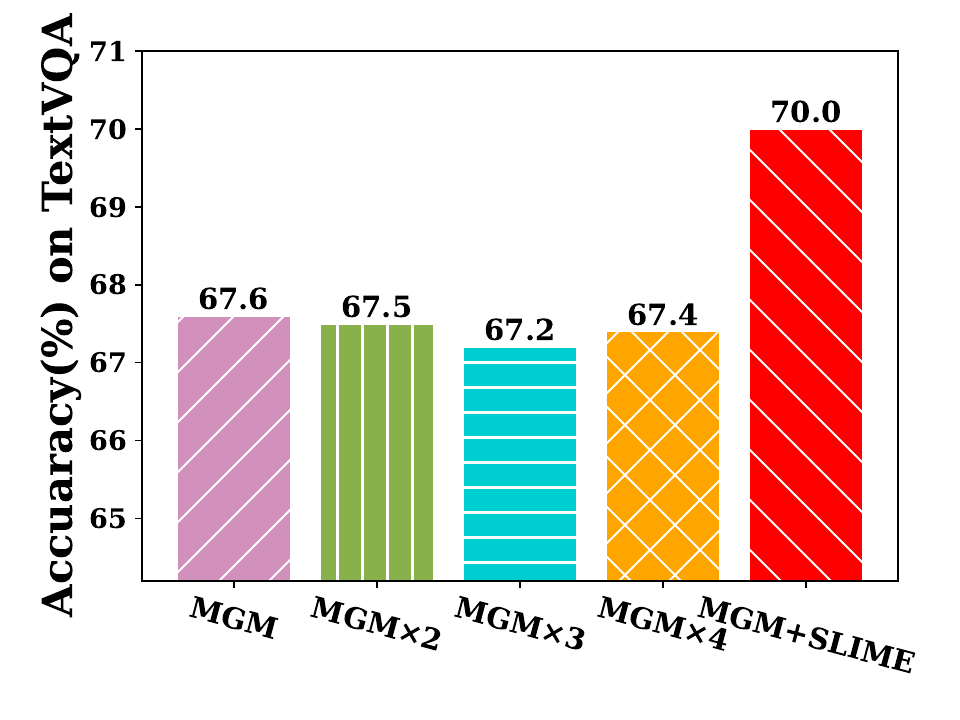}
        \vspace{-0.3cm}
        \caption{Comparision with directly duplicating original tokens many times.}
        \label{fig:copy_vs_merge}
    \end{minipage}
    \hspace{0.015\textwidth} 
    \begin{minipage}{0.31\textwidth}
        \centering
        \includegraphics[width=0.9\linewidth]{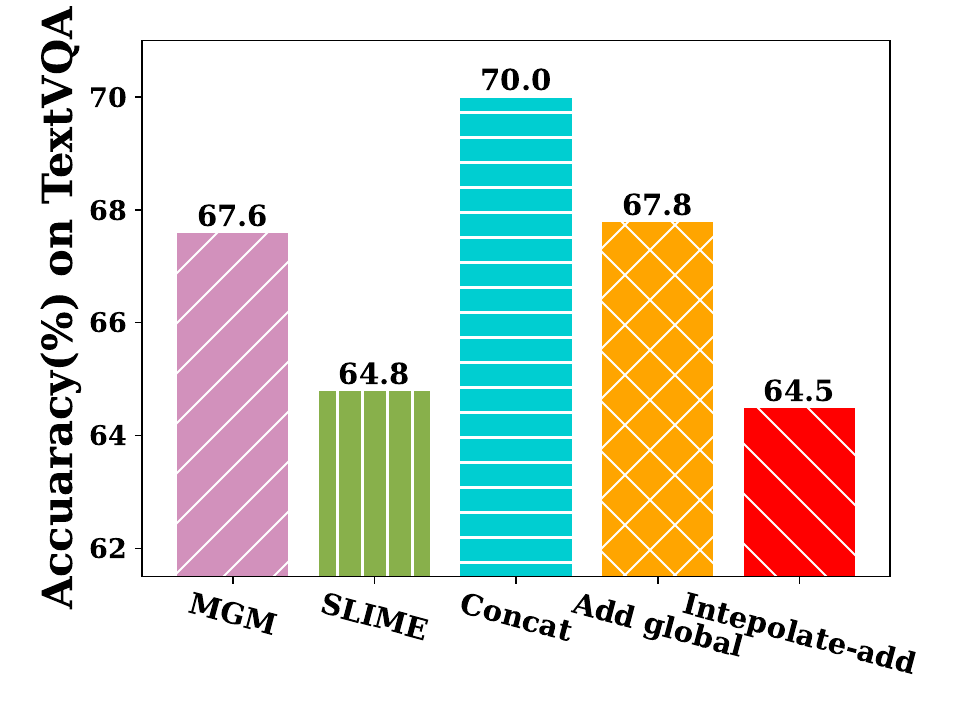}
        \vspace{-0.3cm}
        \caption{Comparisons of concatenating vs. adding different vision tokens.}
        \label{fig:concat_vs_add}
    \end{minipage}
    \hfill
    \begin{minipage}{0.31\textwidth}
        \centering
        \captionsetup{type=table} 
        \small
        \renewcommand{\arraystretch}{1.11}
        \renewcommand{\tabcolsep}{0.1pt}
        \caption{Comparasions of using different model merging methods.}
        \resizebox{\textwidth}{!}{
        \begin{tabular}{l|ccc}
        \toprule
        \textbf{Merging Method} & \textbf{TextVQA} & \textbf{MME\(^P\)} & \textbf{MME\(^E\)} \\
        \midrule
        Task-Arch~\citep{ilharco2022editing} & \textbf{70.0} & \textbf{1645} & \textbf{372} \\
        Average~\citep{choshen2022fusing} & 70.0 & 1632 & 360 \\
        Ties-Merging~\citep{yadav2023resolving} & 69.4 & 1626 & 352 \\
        DARE~\citep{yu2024language} & 69.5 & 1632 & 352 \\
        SLERP~\citep{shoemake1985animating} & 70.0 & 1642 & 368 \\
        \bottomrule
        \end{tabular}
        }
        \label{tab:merge_methods}
    \end{minipage}%
    \vspace{-10pt}
\end{figure*}

\noindent\textbf{Influence of different merge methods.} To evaluate the impact of different model merging strategies on our approach, we apply various methods for merging language models, as presented in Table~\ref{tab:merge_methods}. Regardless of the strategy employed, our method consistently exhibits strong integration performance, surpassing that of the individual models. Among the approaches tested, Task-Arch~\citep{ilharco2022editing} achieved the best results, leading us to adopt it ultimately.

        

\par

\noindent\textbf{Effectiveness of integrating different vision encoders.} To further assess the effectiveness of integrating different vision encoders, we directly duplicate the vision encoder features of MGM-8B multiple times ($\times2$, $\times3$, and $\times4$) and compare the performance of integrating MGM-8B with SLIME-8B. As shown in Figure~\ref{fig:copy_vs_merge}, simply duplicating the visual tokens of MGM-8B, despite providing the language model with more visual token inputs, does not result in performance improvement. This is because no additional visual information is introduced, underscoring the effectiveness of integrating vision encoders from different MLLMs.

\par


\begin{table}[htbp]
\caption{Further evaluation using GPT. We report the average score evaluated by GPT-4.}
\label{tab:gpt_eval}
\centering
\small
\renewcommand{\tabcolsep}{8.0pt}
\resizebox{\columnwidth}{!}{
\begin{tabular}{c|c|c|c}
\toprule
\textbf{Method}        & MGM-8B & SLIME-8B & \cellcolor{gray!20}MGM-SLIME-8B  \\ 
\midrule
\textbf{Average Score} & 7.76   & 7.71     & \textbf{\cellcolor{gray!20} 8.43}(\textcolor{blue}{+8.6\%}) \\ 
\bottomrule
\end{tabular}
}
\end{table}

\noindent\textbf{Further exploration of the enhancement of visual information richness.} Existing visual question-answering datasets primarily focus on querying specific details. To further validate that our method enables models to capture richer visual information, we construct a new dataset by randomly sampling 100 images from the COCO~\citep{lin2014microsoft} dataset. For each image, the text prompt is: ``Please describe this image in as much detail as possible." We use the GPT-4o API to score the model outputs based on the level of detail and accuracy (on a scale of 1-10), inputting both the image and the predictions from individual models and~\methodname. As shown in Table~\ref{tab:gpt_eval}, our method, which integrates vision encoders from different multimodal models, generates more detailed and accurate image descriptions. Compared to the individual models, our approach achieves a relative improvement of 8.6\%, demonstrating that~\methodname~can efficiently leverage perception capabilities of each individual MLLMs and provide more detailed answers. Additional details of evaluation and visualizations of example responses are provided in appendix.

\noindent\textbf{Discussions of concatenating and adding vision tokens.} The length of visual tokens in existing MLLMs varies due to differences in vision encoders, preprocessing techniques, and other factors, making direct fusion through summation challenging. To address this, VisionFuse concatenates features from different vision encoders, effectively handling inconsistent visual token lengths and providing a more flexible and adaptable aggregation method. To further explore the differences between adding and concatenating visual tokens, we conduct experiments using MGM-8B and SLIME-8B. Notably, SLIME captures additional local features through local patches, while its global features share the same length as MGM's. We use two feature summation methods: (A) Add Global, averaging the global features from both models, and (B) Interpolate-Add, where the MGM-8B token sequence is interpolated to match SLIME-8B's length before averaging. Comparison on TextVQA (Figure~\ref{fig:concat_vs_add}) shows that both addition methods underperform concatenation due to information loss from simple averaging. Developing a more effective strategy for fusing visual tokens across different MLLMs is left for future research.



\section{Conclusion}

This paper investigates the differences in visual perception capabilities among various MLLMs and proposes a 
new paradigm
for efficiently enhancing the perceptual abilities of MLLMs based on some novel empirical insights. The approach offers a simple yet effective MLLM integration strategy that requires no additional training for re-aligning the vision encoders and LLM, leveraging the distinct visual perception strengths of different MLLMs to improve performance on multimodal tasks. Overall, \methodname~significantly enhances the perceptual abilities of individual MLLMs with minimal additional deployment overhead. 

\noindent\textbf{Limitations and future work.} Our current analysis focuses on two MLLMs. However, when attempting to integrate more MLLMs, a direct concatenation of visual sequences results in excessively long visual token sequences, causing discrepancies that the sequence length is much longer than that in the training phase and a subsequent decline in performance. Detailed results are provided in appendix. Future work will explore methods for efficiently incorporating additional MLLMs while reducing the length of visual tokens, such as employing token 
merging
strategies across different MLLMs or utilizing rapid fine-tuning techniques to adapt to longer input sequences.


{
    \small
    \bibliographystyle{ieeenat_fullname}
    \bibliography{main}

\begin{thebibliography}{56}
\providecommand{\natexlab}[1]{#1}
\providecommand{\url}[1]{\texttt{#1}}
\expandafter\ifx\csname urlstyle\endcsname\relax
  \providecommand{\doi}[1]{doi: #1}\else
  \providecommand{\doi}{doi: \begingroup \urlstyle{rm}\Url}\fi

\bibitem[AI@Meta(2024)]{llama3modelcard}
AI@Meta.
\newblock Llama 3 model card.
\newblock 2024.

\bibitem[Alayrac et~al.(2022)Alayrac, Donahue, Luc, Miech, Barr, Hasson, Lenc, Mensch, Millican, Reynolds, et~al.]{alayrac2022flamingo}
Jean-Baptiste Alayrac, Jeff Donahue, Pauline Luc, Antoine Miech, Iain Barr, Yana Hasson, Karel Lenc, Arthur Mensch, Katherine Millican, Malcolm Reynolds, et~al.
\newblock Flamingo: a visual language model for few-shot learning.
\newblock \emph{NeurIPS}, 35:\penalty0 23716--23736, 2022.

\bibitem[Bai et~al.(2023)Bai, Bai, Yang, Wang, Tan, Wang, Lin, Zhou, and Zhou]{bai2023qwen}
Jinze Bai, Shuai Bai, Shusheng Yang, Shijie Wang, Sinan Tan, Peng Wang, Junyang Lin, Chang Zhou, and Jingren Zhou.
\newblock Qwen-vl: A frontier large vision-language model with versatile abilities.
\newblock \emph{arXiv preprint arXiv:2308.12966}, 2023.

\bibitem[Bolya et~al.(2022)Bolya, Fu, Dai, Zhang, Feichtenhofer, and Hoffman]{bolya2022tome}
Daniel Bolya, Cheng-Yang Fu, Xiaoliang Dai, Peizhao Zhang, Christoph Feichtenhofer, and Judy Hoffman.
\newblock Token merging: Your vit but faster.
\newblock \emph{arXiv preprint arXiv:2210.09461}, 2022.

\bibitem[Cha et~al.(2024)Cha, Kang, Mun, and Roh]{cha2024honeybee}
Junbum Cha, Wooyoung Kang, Jonghwan Mun, and Byungseok Roh.
\newblock Honeybee: Locality-enhanced projector for multimodal llm.
\newblock In \emph{CVPR}, pages 13817--13827, 2024.

\bibitem[Chen et~al.(2023)Chen, Zhang, Zeng, Zhang, Zhu, and Zhao]{chen2023shikra}
Keqin Chen, Zhao Zhang, Weili Zeng, Richong Zhang, Feng Zhu, and Rui Zhao.
\newblock Shikra: Unleashing multimodal llm's referential dialogue magic.
\newblock \emph{arXiv preprint arXiv:2306.15195}, 2023.

\bibitem[Chen et~al.(2024)Chen, Zhao, Liu, Bai, Lin, Zhou, and Chang]{chen2024fastv}
Liang Chen, Haozhe Zhao, Tianyu Liu, Shuai Bai, Junyang Lin, Chang Zhou, and Baobao Chang.
\newblock An image is worth 1/2 tokens after layer 2: Plug-and-play inference acceleration for large vision-language models, 2024.

\bibitem[Chiang et~al.(2023)Chiang, Li, Lin, Sheng, Wu, Zhang, Zheng, Zhuang, Zhuang, Gonzalez, et~al.]{chiang2023vicuna}
Wei-Lin Chiang, Zhuohan Li, Zi Lin, Ying Sheng, Zhanghao Wu, Hao Zhang, Lianmin Zheng, Siyuan Zhuang, Yonghao Zhuang, Joseph~E Gonzalez, et~al.
\newblock Vicuna: An open-source chatbot impressing gpt-4 with 90\%* chatgpt quality, march 2023.
\newblock 3\penalty0 (5), 2023.

\bibitem[Choshen et~al.(2022)Choshen, Venezian, Slonim, and Katz]{choshen2022fusing}
Leshem Choshen, Elad Venezian, Noam Slonim, and Yoav Katz.
\newblock Fusing finetuned models for better pretraining.
\newblock \emph{arXiv preprint arXiv:2204.03044}, 2022.

\bibitem[Chu et~al.(2023)Chu, Qiao, Lin, Xu, Yang, Hu, Wei, Zhang, Zhang, Wei, et~al.]{chu2023mobilevlm}
Xiangxiang Chu, Limeng Qiao, Xinyang Lin, Shuang Xu, Yang Yang, Yiming Hu, Fei Wei, Xinyu Zhang, Bo Zhang, Xiaolin Wei, et~al.
\newblock Mobilevlm: A fast, reproducible and strong vision language assistant for mobile devices.
\newblock \emph{arXiv preprint arXiv:2312.16886}, 2023.

\bibitem[Dai et~al.(2023)Dai, Li, Li, Tiong, Zhao, Wang, Li, Fung, and Hoi]{dai2023instructblip}
Wenliang Dai, Junnan Li, Dongxu Li, Anthony Meng~Huat Tiong, Junqi Zhao, Weisheng Wang, Boyang Li, Pascale Fung, and Steven Hoi.
\newblock Instructblip: Towards general-purpose vision-language models with instruction tuning, 2023.

\bibitem[Freitag et~al.(2023)Freitag, Ghorbani, and Fernandes]{freitag2023epsilon}
Markus Freitag, Behrooz Ghorbani, and Patrick Fernandes.
\newblock Epsilon sampling rocks: Investigating sampling strategies for minimum bayes risk decoding for machine translation.
\newblock \emph{arXiv preprint arXiv:2305.09860}, 2023.

\bibitem[Fu et~al.(2024)Fu, Chen, Shen, Qin, Zhang, Lin, Yang, Zheng, Li, Sun, Wu, and Ji]{fu2024mmebenchmark}
Chaoyou Fu, Peixian Chen, Yunhang Shen, Yulei Qin, Mengdan Zhang, Xu Lin, Jinrui Yang, Xiawu Zheng, Ke Li, Xing Sun, Yunsheng Wu, and Rongrong Ji.
\newblock Mme: A comprehensive evaluation benchmark for multimodal large language models, 2024.

\bibitem[Ge et~al.(2024)Ge, Cheng, Wang, Yuan, Gao, Song, Song, Huang, and Zheng]{ge2024convllava}
Chunjiang Ge, Sijie Cheng, Ziming Wang, Jiale Yuan, Yuan Gao, Jun Song, Shiji Song, Gao Huang, and Bo Zheng.
\newblock Convllava: Hierarchical backbones as visual encoder for large multimodal models.
\newblock \emph{arXiv preprint arXiv:2405.15738}, 2024.

\bibitem[Goyal et~al.(2017{\natexlab{a}})Goyal, Khot, Summers-Stay, Batra, and Parikh]{goyal2017making}
Yash Goyal, Tejas Khot, Douglas Summers-Stay, Dhruv Batra, and Devi Parikh.
\newblock Making the v in vqa matter: Elevating the role of image understanding in visual question answering.
\newblock In \emph{CVPR}, pages 6904--6913, 2017{\natexlab{a}}.

\bibitem[Goyal et~al.(2017{\natexlab{b}})Goyal, Khot, Summers-Stay, Batra, and Parikh]{goyal2017vqav2}
Yash Goyal, Tejas Khot, Douglas Summers-Stay, Dhruv Batra, and Devi Parikh.
\newblock Making the v in vqa matter: Elevating the role of image understanding in visual question answering.
\newblock In \emph{CVPR}, pages 6904--6913, 2017{\natexlab{b}}.

\bibitem[Gurari et~al.(2018)Gurari, Li, Stangl, Guo, Lin, Grauman, Luo, and Bigham]{gurari2018vizwiz}
Danna Gurari, Qing Li, Abigale~J Stangl, Anhong Guo, Chi Lin, Kristen Grauman, Jiebo Luo, and Jeffrey~P Bigham.
\newblock Vizwiz grand challenge: Answering visual questions from blind people.
\newblock In \emph{CVPR}, pages 3608--3617, 2018.

\bibitem[Han et~al.(2024)Han, Gong, Zhang, Wang, Zhang, Lin, Qiao, Gao, and Yue]{han2024onellm}
Jiaming Han, Kaixiong Gong, Yiyuan Zhang, Jiaqi Wang, Kaipeng Zhang, Dahua Lin, Yu Qiao, Peng Gao, and Xiangyu Yue.
\newblock Onellm: One framework to align all modalities with language.
\newblock In \emph{CVPR}, pages 26584--26595, 2024.

\bibitem[Ilharco et~al.(2022)Ilharco, Ribeiro, Wortsman, Gururangan, Schmidt, Hajishirzi, and Farhadi]{ilharco2022editing}
Gabriel Ilharco, Marco~Tulio Ribeiro, Mitchell Wortsman, Suchin Gururangan, Ludwig Schmidt, Hannaneh Hajishirzi, and Ali Farhadi.
\newblock Editing models with task arithmetic.
\newblock In \emph{ICLR}, 2022.

\bibitem[Ilharco et~al.(2023)Ilharco, Ribeiro, Wortsman, Schmidt, Hajishirzi, and Farhadi]{DBLP:conf/iclr/IlharcoRWSHF23}
Gabriel Ilharco, Marco~T{\'{u}}lio Ribeiro, Mitchell Wortsman, Ludwig Schmidt, Hannaneh Hajishirzi, and Ali Farhadi.
\newblock Editing models with task arithmetic.
\newblock In \emph{ICLR}. OpenReview.net, 2023.

\bibitem[Jiang et~al.(2023)Jiang, Ren, and Lin]{jiang2023llm}
Dongfu Jiang, Xiang Ren, and Bill~Yuchen Lin.
\newblock Llm-blender: Ensembling large language models with pairwise ranking and generative fusion.
\newblock \emph{arXiv preprint arXiv:2306.02561}, 2023.

\bibitem[Jin et~al.(2022)Jin, Ren, Preotiuc-Pietro, and Cheng]{jin2022dataless}
Xisen Jin, Xiang Ren, Daniel Preotiuc-Pietro, and Pengxiang Cheng.
\newblock Dataless knowledge fusion by merging weights of language models.
\newblock In \emph{ICLR}, 2022.

\bibitem[Kembhavi et~al.(2016)Kembhavi, Salvato, Kolve, Seo, Hajishirzi, and Farhadi]{kembhavi2016diagram}
Aniruddha Kembhavi, Mike Salvato, Eric Kolve, Minjoon Seo, Hannaneh Hajishirzi, and Ali Farhadi.
\newblock A diagram is worth a dozen images.
\newblock In \emph{ECCV}, pages 235--251. Springer, 2016.

\bibitem[Laurencon et~al.(2023)Laurencon, van Strien, Bekman, Tronchon, Saulnier, Wang, Karamcheti, Singh, Pistilli, Jernite, et~al.]{laurencon2023introducing}
Hugo Laurencon, Daniel van Strien, Stas Bekman, Leo Tronchon, Lucile Saulnier, Thomas Wang, Siddharth Karamcheti, Amanpreet Singh, Giada Pistilli, Yacine Jernite, et~al.
\newblock Introducing idefics: An open reproduction of state-of-the-art visual language model, 2023.

\bibitem[Li et~al.(2023{\natexlab{a}})Li, Li, Savarese, and Hoi]{li2023blip}
Junnan Li, Dongxu Li, Silvio Savarese, and Steven Hoi.
\newblock Blip-2: Bootstrapping language-image pre-training with frozen image encoders and large language models.
\newblock In \emph{ICML}, pages 19730--19742. PMLR, 2023{\natexlab{a}}.

\bibitem[Li et~al.(2023{\natexlab{b}})Li, Wang, and Jia]{li2023llama}
Yanwei Li, Chengyao Wang, and Jiaya Jia.
\newblock Llama-vid: An image is worth 2 tokens in large language models.
\newblock \emph{arXiv preprint arXiv:2311.17043}, 2023{\natexlab{b}}.

\bibitem[Li et~al.(2024)Li, Zhang, Wang, Zhong, Chen, Chu, Liu, and Jia]{li2024mini}
Yanwei Li, Yuechen Zhang, Chengyao Wang, Zhisheng Zhong, Yixin Chen, Ruihang Chu, Shaoteng Liu, and Jiaya Jia.
\newblock Mini-gemini: Mining the potential of multi-modality vision language models.
\newblock \emph{arXiv preprint arXiv:2403.18814}, 2024.

\bibitem[Lin et~al.(2023)Lin, Zhu, Ye, Ning, Jin, and Yuan]{lin2023video}
Bin Lin, Bin Zhu, Yang Ye, Munan Ning, Peng Jin, and Li Yuan.
\newblock Video-llava: Learning united visual representation by alignment before projection.
\newblock \emph{arXiv preprint arXiv:2311.10122}, 2023.

\bibitem[Lin et~al.(2024)Lin, Yin, Ping, Molchanov, Shoeybi, and Han]{lin2024vila}
Ji Lin, Hongxu Yin, Wei Ping, Pavlo Molchanov, Mohammad Shoeybi, and Song Han.
\newblock Vila: On pre-training for visual language models.
\newblock In \emph{CVPR}, pages 26689--26699, 2024.

\bibitem[Lin et~al.(2014)Lin, Maire, Belongie, Hays, Perona, Ramanan, Doll{\'a}r, and Zitnick]{lin2014microsoft}
Tsung-Yi Lin, Michael Maire, Serge Belongie, James Hays, Pietro Perona, Deva Ramanan, Piotr Doll{\'a}r, and C~Lawrence Zitnick.
\newblock Microsoft coco: Common objects in context.
\newblock In \emph{ECCV}, pages 740--755. Springer, 2014.

\bibitem[Liu et~al.(2023{\natexlab{a}})Liu, Lin, Li, Wang, Yacoob, and Wang]{liu2023mitigating}
Fuxiao Liu, Kevin Lin, Linjie Li, Jianfeng Wang, Yaser Yacoob, and Lijuan Wang.
\newblock Mitigating hallucination in large multi-modal models via robust instruction tuning.
\newblock In \emph{CVPR}, 2023{\natexlab{a}}.

\bibitem[Liu et~al.(2024{\natexlab{a}})Liu, Li, Li, and Lee]{liu2024improved}
Haotian Liu, Chunyuan Li, Yuheng Li, and Yong~Jae Lee.
\newblock Improved baselines with visual instruction tuning.
\newblock In \emph{CVPR}, pages 26296--26306, 2024{\natexlab{a}}.

\bibitem[Liu et~al.(2024{\natexlab{b}})Liu, Li, Li, Li, Zhang, Shen, and Lee]{liu2024llavanext}
Haotian Liu, Chunyuan Li, Yuheng Li, Bo Li, Yuanhan Zhang, Sheng Shen, and Yong~Jae Lee.
\newblock Llava-next: Improved reasoning, ocr, and world knowledge, 2024{\natexlab{b}}.

\bibitem[Liu et~al.(2024{\natexlab{c}})Liu, Li, Wu, and Lee]{liu2024visual}
Haotian Liu, Chunyuan Li, Qingyang Wu, and Yong~Jae Lee.
\newblock Visual instruction tuning.
\newblock \emph{NeurIPS}, 36, 2024{\natexlab{c}}.

\bibitem[Liu et~al.(2024{\natexlab{d}})Liu, Xiao, Li, Lee, Han, Dao, and Cai]{liu2024bitdelta}
James Liu, Guangxuan Xiao, Kai Li, Jason~D Lee, Song Han, Tri Dao, and Tianle Cai.
\newblock Bitdelta: Your fine-tune may only be worth one bit.
\newblock \emph{arXiv preprint arXiv:2402.10193}, 2024{\natexlab{d}}.

\bibitem[Liu et~al.(2023{\natexlab{b}})Liu, Duan, Zhang, Li, Zhang, Zhao, Yuan, Wang, He, Liu, et~al.]{liu2023mmbench}
Yuan Liu, Haodong Duan, Yuanhan Zhang, Bo Li, Songyang Zhang, Wangbo Zhao, Yike Yuan, Jiaqi Wang, Conghui He, Ziwei Liu, et~al.
\newblock Mmbench: Is your multi-modal model an all-around player?
\newblock \emph{arXiv preprint arXiv:2307.06281}, 2023{\natexlab{b}}.

\bibitem[Liu et~al.(2023{\natexlab{c}})Liu, Li, Yang, Li, Yin, Liu, Jin, and Bai]{liu2023hidden}
Yuliang Liu, Zhang Li, Biao Yang, Chunyuan Li, Xucheng Yin, Cheng-lin Liu, Lianwen Jin, and Xiang Bai.
\newblock On the hidden mystery of ocr in large multimodal models.
\newblock \emph{arXiv preprint arXiv:2305.07895}, 2023{\natexlab{c}}.

\bibitem[Lu et~al.(2023)Lu, Li, Liu, Yang, Gao, and Shen]{lu2023empirical}
Yadong Lu, Chunyuan Li, Haotian Liu, Jianwei Yang, Jianfeng Gao, and Yelong Shen.
\newblock An empirical study of scaling instruct-tuned large multimodal models.
\newblock \emph{arXiv preprint arXiv:2309.09958}, 2023.

\bibitem[Radford et~al.(2021)Radford, Kim, Hallacy, Ramesh, Goh, Agarwal, Sastry, Askell, Mishkin, Clark, et~al.]{radford2021clip}
Alec Radford, Jong~Wook Kim, Chris Hallacy, Aditya Ramesh, Gabriel Goh, Sandhini Agarwal, Girish Sastry, Amanda Askell, Pamela Mishkin, Jack Clark, et~al.
\newblock Learning transferable visual models from natural language supervision.
\newblock In \emph{ICML}, pages 8748--8763. PMLR, 2021.

\bibitem[Shi et~al.(2024)Shi, Liu, Wang, Liao, Radhakrishnan, Huang, Yin, Sapra, Yacoob, Shi, et~al.]{shi2024eagle}
Min Shi, Fuxiao Liu, Shihao Wang, Shijia Liao, Subhashree Radhakrishnan, De-An Huang, Hongxu Yin, Karan Sapra, Yaser Yacoob, Humphrey Shi, et~al.
\newblock Eagle: Exploring the design space for multimodal llms with mixture of encoders.
\newblock \emph{arXiv preprint arXiv:2408.15998}, 2024.

\bibitem[Shoemake(1985)]{shoemake1985animating}
Ken Shoemake.
\newblock Animating rotation with quaternion curves.
\newblock In \emph{Proceedings of the 12th annual conference on Computer graphics and interactive techniques}, pages 245--254, 1985.

\bibitem[Singh et~al.(2019)Singh, Natarajan, Shah, Jiang, Chen, Batra, Parikh, and Rohrbach]{singh2019towards}
Amanpreet Singh, Vivek Natarajan, Meet Shah, Yu Jiang, Xinlei Chen, Dhruv Batra, Devi Parikh, and Marcus Rohrbach.
\newblock Towards vqa models that can read.
\newblock In \emph{CVPR}, pages 8317--8326, 2019.

\bibitem[Van~der Maaten and Hinton(2008)]{van2008tsne}
Laurens Van~der Maaten and Geoffrey Hinton.
\newblock Visualizing data using t-sne.
\newblock \emph{Journal of machine learning research}, 9\penalty0 (11), 2008.

\bibitem[Wan et~al.(2024)Wan, Huang, Cai, Quan, Bi, and Shi]{wan2024knowledge}
Fanqi Wan, Xinting Huang, Deng Cai, Xiaojun Quan, Wei Bi, and Shuming Shi.
\newblock Knowledge fusion of large language models.
\newblock \emph{arXiv preprint arXiv:2401.10491}, 2024.

\bibitem[{xAI}(2024)]{xai2024grok}
{xAI}.
\newblock {Grok-1.5 Vision Preview}, 2024.
\newblock Accessed: November 2024.

\bibitem[Xu et~al.(2024)Xu, Zhao, Zhou, Lin, Ng, and Feng]{xu2024pllava}
Lin Xu, Yilin Zhao, Daquan Zhou, Zhijie Lin, See~Kiong Ng, and Jiashi Feng.
\newblock Pllava: Parameter-free llava extension from images to videos for video dense captioning.
\newblock \emph{arXiv preprint arXiv:2404.16994}, 2024.

\bibitem[Yadav et~al.(2023)Yadav, Tam, Choshen, Raffel, and Bansal]{yadav2023resolving}
Prateek Yadav, Derek Tam, Leshem Choshen, Colin Raffel, and Mohit Bansal.
\newblock Resolving interference when merging models.
\newblock \emph{arXiv preprint arXiv:2306.01708}, 2023.

\bibitem[Yang et~al.(2024)Yang, Zhai, You, Yuan, Yang, and Xu]{yang2024law}
Shijia Yang, Bohan Zhai, Quanzeng You, Jianbo Yuan, Hongxia Yang, and Chenfeng Xu.
\newblock Law of vision representation in mllms.
\newblock \emph{arXiv preprint arXiv:2408.16357}, 2024.

\bibitem[Ye et~al.(2023)Ye, Xu, Xu, Ye, Yan, Zhou, Wang, Hu, Shi, Shi, et~al.]{ye2023mplug}
Qinghao Ye, Haiyang Xu, Guohai Xu, Jiabo Ye, Ming Yan, Yiyang Zhou, Junyang Wang, Anwen Hu, Pengcheng Shi, Yaya Shi, et~al.
\newblock mplug-owl: Modularization empowers large language models with multimodality.
\newblock \emph{arXiv preprint arXiv:2304.14178}, 2023.

\bibitem[Yin et~al.(2023)Yin, Fu, Zhao, Li, Sun, Xu, and Chen]{yin2023survey}
Shukang Yin, Chaoyou Fu, Sirui Zhao, Ke Li, Xing Sun, Tong Xu, and Enhong Chen.
\newblock A survey on multimodal large language models.
\newblock \emph{arXiv preprint arXiv:2306.13549}, 2023.

\bibitem[Yu et~al.(2024)Yu, Yu, Yu, Huang, and Li]{yu2024language}
Le Yu, Bowen Yu, Haiyang Yu, Fei Huang, and Yongbin Li.
\newblock Language models are super mario: Absorbing abilities from homologous models as a free lunch.
\newblock In \emph{ICML}, 2024.

\bibitem[Yue et~al.(2024)Yue, Ni, Zhang, Zheng, Liu, Zhang, Stevens, Jiang, Ren, Sun, et~al.]{yue2024mmmu}
Xiang Yue, Yuansheng Ni, Kai Zhang, Tianyu Zheng, Ruoqi Liu, Ge Zhang, Samuel Stevens, Dongfu Jiang, Weiming Ren, Yuxuan Sun, et~al.
\newblock Mmmu: A massive multi-discipline multimodal understanding and reasoning benchmark for expert agi.
\newblock In \emph{CVPR}, pages 9556--9567, 2024.

\bibitem[Zhang et~al.(2023)Zhang, Zhang, Gu, Zhou, Lipka, Yang, and Sun]{zhang2023llavar}
Yanzhe Zhang, Ruiyi Zhang, Jiuxiang Gu, Yufan Zhou, Nedim Lipka, Diyi Yang, and Tong Sun.
\newblock Llavar: Enhanced visual instruction tuning for text-rich image understanding.
\newblock \emph{arXiv preprint arXiv:2306.17107}, 2023.

\bibitem[Zhang et~al.(2024)Zhang, Wen, Fu, Wang, Zhang, Wang, and Jin]{zhang2024beyond}
Yi-Fan Zhang, Qingsong Wen, Chaoyou Fu, Xue Wang, Zhang Zhang, Liang Wang, and Rong Jin.
\newblock Beyond llava-hd: Diving into high-resolution large multimodal models.
\newblock \emph{arXiv preprint arXiv:2406.08487}, 2024.

\bibitem[Zhao et~al.(2023)Zhao, Wu, He, and Huang]{zhao2023svit}
Bo Zhao, Boya Wu, Muyang He, and Tiejun Huang.
\newblock Svit: Scaling up visual instruction tuning.
\newblock \emph{arXiv preprint arXiv:2307.04087}, 2023.

\bibitem[Zhu et~al.(2023)Zhu, Chen, Shen, Li, and Elhoseiny]{zhu2023minigpt}
Deyao Zhu, Jun Chen, Xiaoqian Shen, Xiang Li, and Mohamed Elhoseiny.
\newblock Minigpt-4: Enhancing vision-language understanding with advanced large language models.
\newblock \emph{arXiv preprint arXiv:2304.10592}, 2023.

\end{thebibliography}
}

\clearpage
\appendix
\setcounter{page}{1}

\def\mytitle{
VISIONFUSE: ENHANCING MULTIMODAL LLMS
PERCEPTION FOR FREE
}

\begin{center}
	{
        \Large{Appendix}
	}
\end{center}

\etocdepthtag.toc{mtappendix}
\etocsettagdepth{mtchapter}{none}
\etocsettagdepth{mtappendix}{subsection}

{
    \footnotesize\tableofcontents
}

\section{Details of Experimental Settings}

\subsection{Details of Evaluation using GPT}
\label{sec:details_gpt}

For each example, we include both the image and the outputs from multiple models with the same prompt. The images are provided as URLs in the GPT-4o API. GPT-4o then returns a JSON object containing the scores for all models. The prompt is as follows:

\begin{lstlisting}
Next, I will provide you with descriptions of an image generated by multiple models. Please evaluate these descriptions based on the level of detail and accuracy, and assign a score ranging from 1 to 10. Finally, your output only contains a JSON object, where each item is the model name and its corresponding score.

model A: ......

model B: ......

model C: ......
\end{lstlisting}

\subsection{Searching Details for the Hyper-parameters of Merging Methods}

Table~\ref{tab:hyperparameter-ranges} shows the searching range of the parameters of several merging methods. 

\begin{table}[ht]
\caption{Searched ranges of hyperparameters.}
\label{tab:hyperparameter-ranges}
\centering
\begin{tabular}{p{3.4cm}|p{4.5cm}}
\toprule
\textbf{Method} & \textbf{Search Ranges of Hyperparameters} \\
\midrule
Task Arithmetic~\citep{ilharco2022editing} & Scaling term: [0.1, 0.3, 0.5, 0.7, 0.9, 1.0] \\
\midrule
TIES-Merging~\citep{yadav2023resolving} & Scaling term: [0.1, 0.3, 0.5, 0.7, 0.9, 1.0] \\
& Ratio of retain parameters: [0.1, 0.2, 0.3] \\
\midrule
DARE~\citep{yu2024language} & Scaling term: [0.1, 0.3, 0.5, 0.7, 0.9, 1.0] \\
& Drop rate: [0.1, 0.3, 0.5, 0.7, 0.9] \\
\bottomrule
\end{tabular}

\end{table}

\section{Additional Related Work}

\textbf{Multimodal Large Language Models.} MLLMs integrate vision encoders into large language models, enabling them to handle multimodal tasks. Early models such as Flamingo~\citep{alayrac2022flamingo} encode images and feed them into the attention layers of the language model, while Blip-2~\citep{li2023blip} employs Q-Former to encode images into features, which are then input into the language model. Subsequent works~\citep{liu2024visual,liu2023mitigating,lu2023empirical,zhang2023llavar,zhao2023svit,zhu2023minigpt} enhance the multimodal understanding capabilities of language models through instruction fine-tuning on multimodal datasets. Further research has focused on optimizing encoder designs, extracting richer visual information, and expanding the models to handle additional modalities. For example, Eagle~\citep{shi2024eagle}, Mini-Gemini~\citep{li2024mini}, SLIME~\citep{zhang2024beyond}, and LLaVA-next~\citep{liu2024llavanext} introduce additional vision encoders and employ preprocessing techniques such as cropping and interpolation to handle longer visual input sequences, thereby enriching the visual information available to the language model. Honeybee~\citep{cha2024honeybee} introduces a locality-enhanced visual projector to better bridge pre-trained vision encoders with large language models. Recent works~\citep{lin2023video,xu2024pllava,han2024onellm} also explore enabling language models to understand a wider range of modalities.

\begin{table*}[ht]
\caption{Comparison with leading methods on more model combinations and multimodal benchmarks. Results of~\methodname~are marked in gray. VQA\(^T\): TextVQA; MMB: MMBench; MMB\(^C\): MMBench-Chinese; MMMU\(_{v,t}\): validation and test set of MMMU; MME\(^{P,C}\): Perception and Cognition in MME; RWQA: RealworldQA; OCR: OCRBench. Res. indicates the input resolution. Percentages indicate the rate of improvement compared to the best performance of the baselines.}
\label{tab:supp_main_results}
\centering
\small
\renewcommand{\tabcolsep}{2.0pt}
\begin{tabular}{l|c|c|c|c|c|c|c|c|c|c|c|c|c}
\toprule
\textbf{Method} & \textbf{Res.} & \textbf{VQA\(^T\)} & \textbf{VQAv2} & \textbf{Vizwiz} & \textbf{MME\(^P\)} & \textbf{MME\(^C\)} & \textbf{MMB} & \textbf{MMB\(^C\)} & \textbf{MMMU$_v$} & \textbf{MMMU$_t$} & \textbf{Ai2d} & \textbf{RWQA} & \textbf{OCR} \\ 
\midrule
MobileVLM & 336 & 47.5 & - & - & 1289 & - & 59.6 & - & 26.2 & - & - & - & - \\ 
InstructBLIP & 224 & 50.1 & - & 34.5 & - & - & - & 36.0 & - & - & - & - & - \\ 
InstructBLIP & 336 & 50.7 & - & 33.4 & 1213 & - & - & - & 25.6 & - & - & - & - \\ 
Qwen-VL & 336 & 59.8 & 78.8 & 35.2 & - & - & 66.0 & - & - & - & - & - & - \\ 
Qwen-VL-Chat & 448 & 61.5 & 78.2 & 38.9 & 1488 & - & 68.0 & - & 35.9 & 32.2 & - & - & - \\ 
Shikra & 336 & 52.3 & - & - & - & - & 59.2 & - & - & - & - & - & - \\ 
IDEFICS-80B & 224 & 30.9 & - & - & - & - & 54.5 & - & - & - & - & - & - \\ 
LLaVA-1.5 & 336 & 53.5 & 78.5 & 50.0 & - & - & 66.4 & 58.3 & 31.5 & - & - & - & - \\ 
LLaVA-1.5 & 336 & 63.2 & 80.0 & 53.6 & 1531 & 295 & 65.2 & 63.6 & 36.4 & 33.1 & - & - & - \\ 
LLaVA-HD & 336 & 62.5 & 81.8 & 57.5 & 1500 & - & 68.8 & 61.9 & - & - & - & - & - \\ 
LLaVA-Next-8B & 336 & 64.6 & - & - & 1604 & 318 & 72.1 & - & 40.7 & 37.0 & 71.6 & 60.1 & 49.0 \\
MGM-8x7B & 336 & 69.2 & - & - & 1639 & 379 & 75.6 & - & 41.8 & 37.1 & - & - & - \\ 
\midrule
MGM-7B & 336 & 65.2 & 80.4 & 52.1 & 1523 & 316 & 68.7 & 57.8 & 36.1 & 32.8 & 66.3 & 56.7 & 43.6 \\ 
SliME-7B & 336 & 64.4 & 80.3 & 53.7 & 1544 & 383 & 68.4 & 61.3 & 37.2 & 33.4 & 63.2 & 55.4 & 38.3 \\ 
\rowcolor{gray!20}
 &  & \textbf{66.9} & \textbf{80.7} & \textbf{54.4} & \textbf{1563} & \textbf{394} & \textbf{69.6} & \textbf{62.5} & \textbf{37.8} & \textbf{33.6} & \textbf{67.2} & \textbf{57.6} & \textbf{45.3} \\ 
\rowcolor{gray!20}
\multirow{-2}{*}{MGM-SliME-7B} & \multirow{-2}{*}{336} & \textcolor{blue}{+2.6\%} & \textcolor{blue}{+0.4\%} & \textcolor{blue}{+1.3\%} & \textcolor{blue}{+1.2\%} & \textcolor{blue}{+2.9\%} & \textcolor{blue}{+1.3\%} & \textcolor{blue}{+2.0\%} & \textcolor{blue}{+1.6\%} & \textcolor{blue}{+0.6\%} & \textcolor{blue}{+1.4\%} & \textcolor{blue}{+1.6\%} & \textcolor{blue}{+3.9\%} \\ 
\midrule
MGM-8B & 336 & 67.6 & 81.0 & 50.9 & 1606 & 341 & 68.1 & 62.1 & 38.2 & 36.3 & 72.3 & 56.2 & 46.3 \\ 
SliME-8B & 336 & 64.8 & 80.7 & 53.1 & 1578 & 337 & 73.2 & 69.8 & 40.8 & 37.2 & 70.9 & 60.5 & 44.8 \\ 
\rowcolor{gray!20}
 &  & \textbf{70.0} & \textbf{82.1} & \textbf{60.9} & \textbf{1645} & \textbf{372} & \textbf{73.9} & \textbf{71.9} & \textbf{41.6} & \textbf{38.4} & \textbf{75.2} & \textbf{61.3} & \textbf{49.3} \\ 
\rowcolor{gray!20}
\multirow{-2}{*}{MGM-SliME-8B} & \multirow{-2}{*}{336} & \textcolor{blue}{+3.6\%} & \textcolor{blue}{+1.4\%} & \textcolor{blue}{+14.7\%} & \textcolor{blue}{+2.4\%} & \textcolor{blue}{+9.1\%} & \textcolor{blue}{+1.0\%} & \textcolor{blue}{+3.0\%} & \textcolor{blue}{+2.0\%} & \textcolor{blue}{+3.2\%} & \textcolor{blue}{+4.0\%} & \textcolor{blue}{+1.3\%} & \textcolor{blue}{+6.5\%} \\

\midrule
MGM-HD & 672 & 71.6 & 81.5 & 54.4 & 1532 & 357 & 70.6 & 65.2 & 37.0 & 36.5 & 72.5 & 60.5 & 54.6 \\ 
\rowcolor{gray!20}
 &  & \textbf{72.7} & \textbf{82.2} & \textbf{55.2} & \textbf{1600} & \textbf{364} & \textbf{75.1} & \textbf{70.8} & \textbf{41.8} & \textbf{37.4} & \textbf{73.8} & \textbf{61.3} & \textbf{54.9} \\ 
\rowcolor{gray!20}
\multirow{-2}{*}{MGM-HD-SliME-8B} & \multirow{-2}{*}{672} & \textcolor{blue}{+1.5\%} & \textcolor{blue}{+0.9\%} & \textcolor{blue}{+1.5\%} & \textcolor{blue}{+1.4\%} & \textcolor{blue}{+2.0\%} & \textcolor{blue}{+2.5\%} & \textcolor{blue}{+1.4\%} & \textcolor{blue}{+2.5\%} & \textcolor{blue}{+0.5\%} & \textcolor{blue}{+1.7\%} & \textcolor{blue}{+1.3\%} & \textcolor{blue}{+0.5\%} \\

\midrule

Eagle-8B & 1024 & 77.1 & 84.7 & 67.9 & 1568 & 400 & 75.9 & 70.2 & 38.1 & 35.8 & 75.4 & 66.5 & 61.9 \\ 
\rowcolor{gray!20}
 &  & \textbf{77.6} & \textbf{84.8} & \textbf{68.8} & \textbf{1625} & \textbf{408} & \textbf{78.1} & \textbf{73.2} & \textbf{41.8} & \textbf{39.0} & \textbf{76.7} & \textbf{67.6} & \textbf{63.9} \\ 
 \rowcolor{gray!20}
\multirow{-2}{*}{MGM-Eagle-8B} & \multirow{-2}{*}{1024} & \textcolor{blue}{+0.6\%} & \textcolor{blue}{+0.1\%} & \textcolor{blue}{+1.3\%} & \textcolor{blue}{+1.2\%} & \textcolor{blue}{+2.0\%} & \textcolor{blue}{+2.9\%} & \textcolor{blue}{+4.3\%} & \textcolor{blue}{+8.6\%} & \textcolor{blue}{+7.4\%} & \textcolor{blue}{+6.9\%} & \textcolor{blue}{+1.7\%} & \textcolor{blue}{+3.2\%} \\
\bottomrule
\end{tabular}
\end{table*}

\section{Performance on More Datasets and MLLMs Combination}

In this section, we present the experimental results on various multimodal datasets, including Ai2d~\citep{kembhavi2016diagram}, RealworldQA~\citep{xai2024grok}, and OCRBench~\citep{liu2023hidden}. Furthermore, we explore additional combinations of MLLMs, such as MGM-8B~\citep{li2024mini} and Eagle-8B~\citep{shi2024eagle}.

As shown in Table~\ref{tab:supp_main_results}, the improvement on the Ai2d (knowledge evaluation), OCRBench (text recognition), and RealworldQA (visual-centric) datasets demonstrates that our method effectively integrates the perception capabilities of individual models across diverse tasks, showcasing its generalization ability. Notably, Eagle-8B improves its visual perception by integrating multiple vision encoders through extensive training. In contrast, our method requires no additional training and can leverage external vision encoders to further enhance the performance of Eagle-8B, underscoring the effectiveness of our approach.

\section{Discussions and Limitations}

\label{sec:long_sequence}
\begin{figure*}[t]
    \centering
    \begin{subfigure}[b]{0.32\textwidth}
        \centering
        \includegraphics[width=\textwidth]{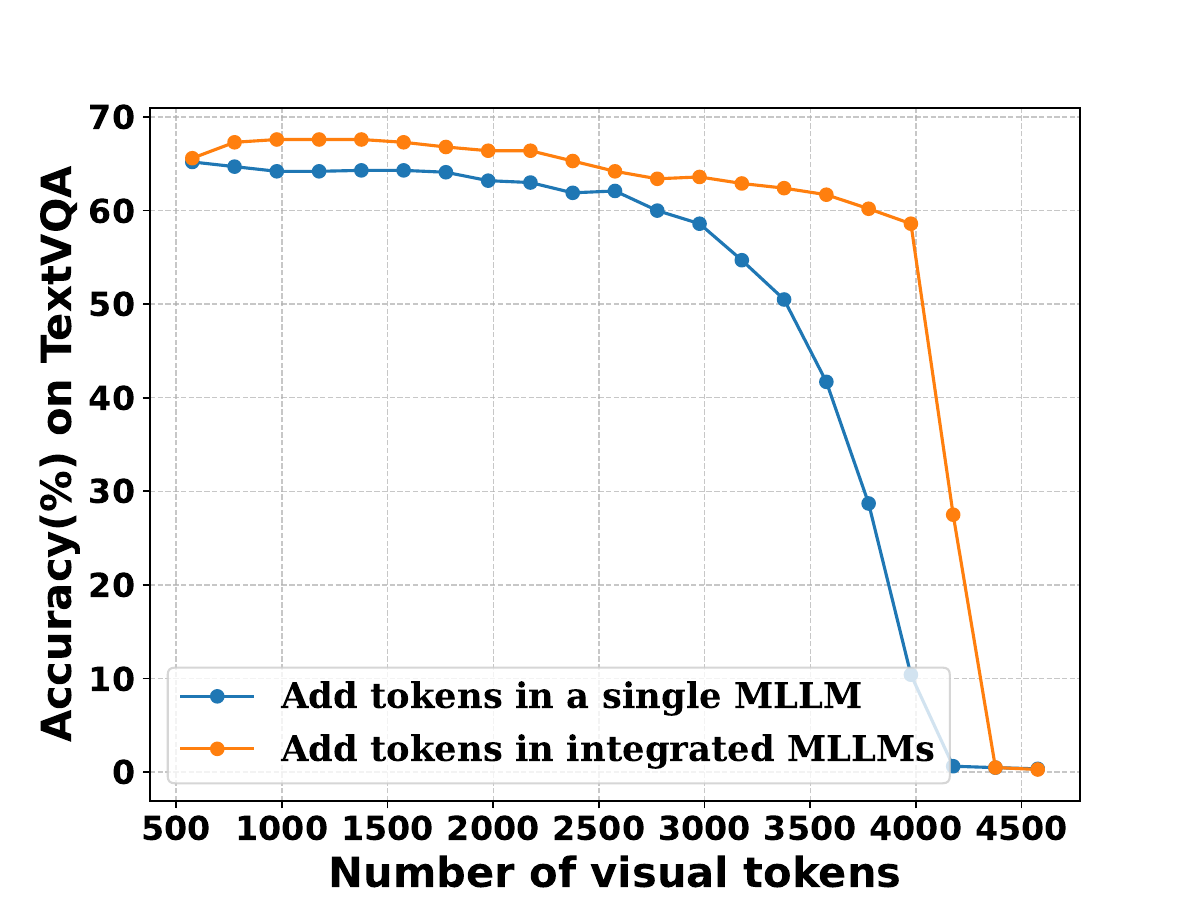}
        \caption{Impact of increasing token length.}
        \label{fig:issue_token_length}
    \end{subfigure}
    \hfill
    \begin{subfigure}[b]{0.327\textwidth}
        \centering
        \includegraphics[width=\textwidth]{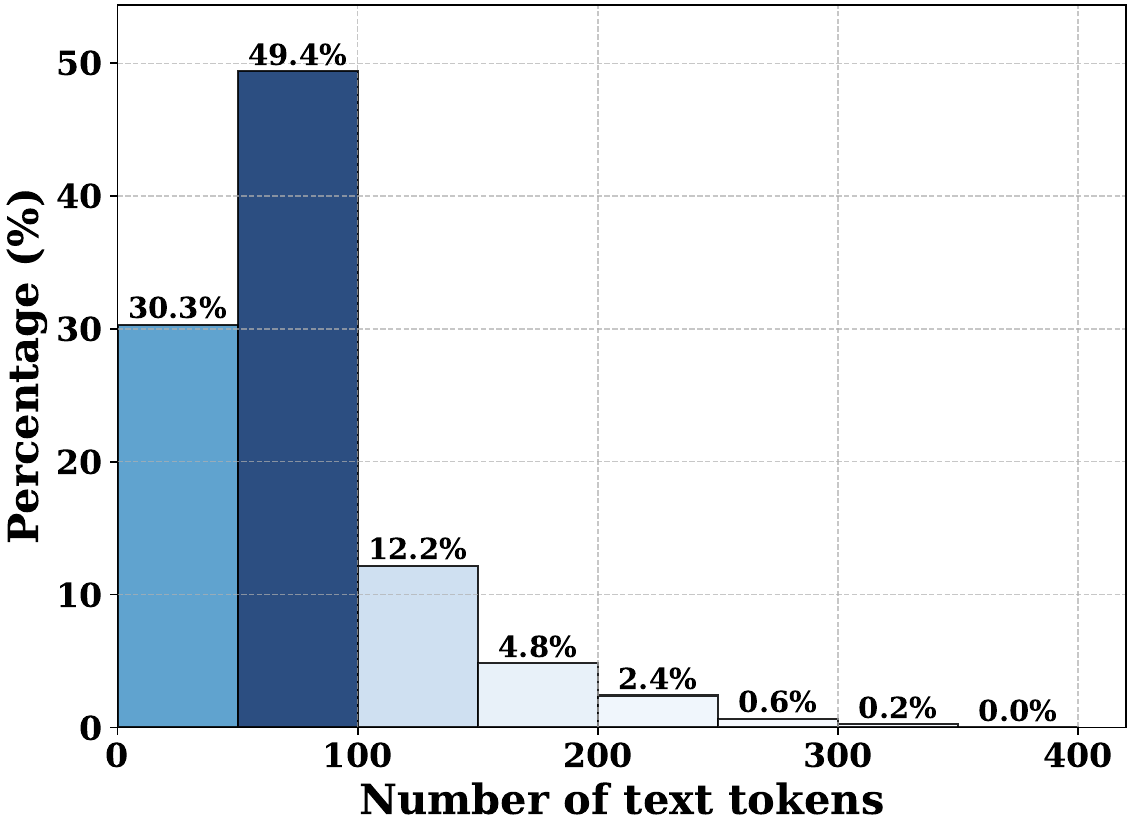}
        \caption{Statistics of text token length in TextVQA.}
        \label{fig:text_token_length}
    \end{subfigure}
    \hfill
    \begin{subfigure}[b]{0.327\textwidth}
        \centering
    \includegraphics[width=\textwidth]{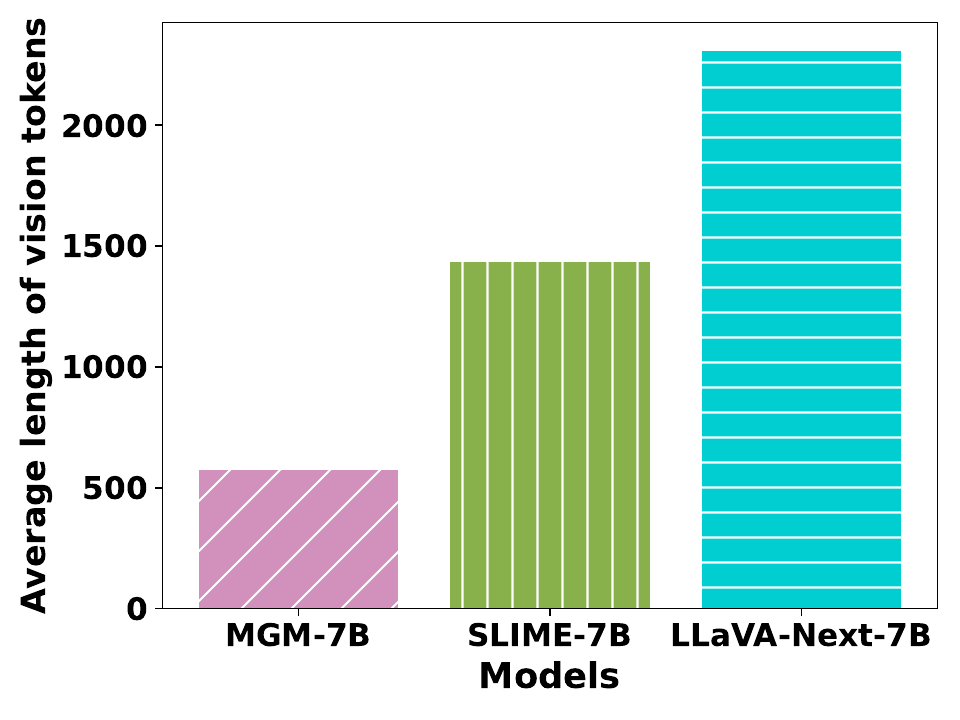}
        \caption{Average length of vision tokens in TextVQA.}
        \label{fig:vision_token_length}
    \end{subfigure}
    \caption{Exploration of the impact of sequence length: (a) demonstrates the significant performance degradation caused by visual sequences that are excessively long and inconsistent with the training phase; (b) provides statistics on text sequence lengths in the TextVQA dataset; and (c) presents the average length of visual sequences during testing on TextVQA for the three models.}
    \label{fig:appendix_c1}
\end{figure*}

\subsection{Impact of the Token Sequence Length}
Although the proposed \methodname~enhances visual perception capabilities by effectively integrating the vision encoders of different models through the concatenation of visual tokens, excessively long token sequences can introduce challenges.
In Section~\ref{sec:long_sequence}, we discuss this issue in detail. In Section~\ref{sec:token_pruning}, we further discuss the underlying solution to alleviate this limitation.


In this part, we investigate the impact of the token sequence length on the model performance.
We progressively increase the number of random visual tokens and show the results of MGM-7B (a single MLLM) and our \methodname~(integration of MGM-7B and SLIME-7B). 
As shown in Figure~\ref{fig:issue_token_length}, a significant decline in model performance is observed when the length of visual tokens exceeds a certain threshold. This decline can be attributed to the mismatch between the shorter token lengths used during training and the longer ones encounter during inference. 

To verify the above conclusion, we also analyze the number of input text and visual tokens for MGM-7B during training and inference on the TextVQA dataset. 
As illustrated in Figure~\ref{fig:text_token_length}, the majority of samples contain between 0 and 100 text tokens. During training, the visual token sequence length for both models is capped at 4096, meaning that the models are not exposed to sequences longer than this during multimodal fine-tuning. This limitation leeds to degraded performance when longer sequences are encountered during testing. As shown in Figure~\ref{fig:vision_token_length}, the total visual token length for MGM-7B, SLIME-7B, and LLaVA-Next-7B exceeds 4096, which surpasses the sequence length encounters during training. 
Consequently, all three MLLMs may not learn how to handle the longer token sequences, resulting in a marked performance decline when encountering these token sequences after the integration of these MLLMs, as illustrated in Table~\ref{tab:main_results_appendix}.

\subsection{Exploration of Integrating more Models through Token Pruning}
\label{sec:token_pruning}
\begin{figure*}[ht]
    \centering
    \begin{subfigure}[b]{0.325\textwidth}
        \centering
        \includegraphics[width=\textwidth]{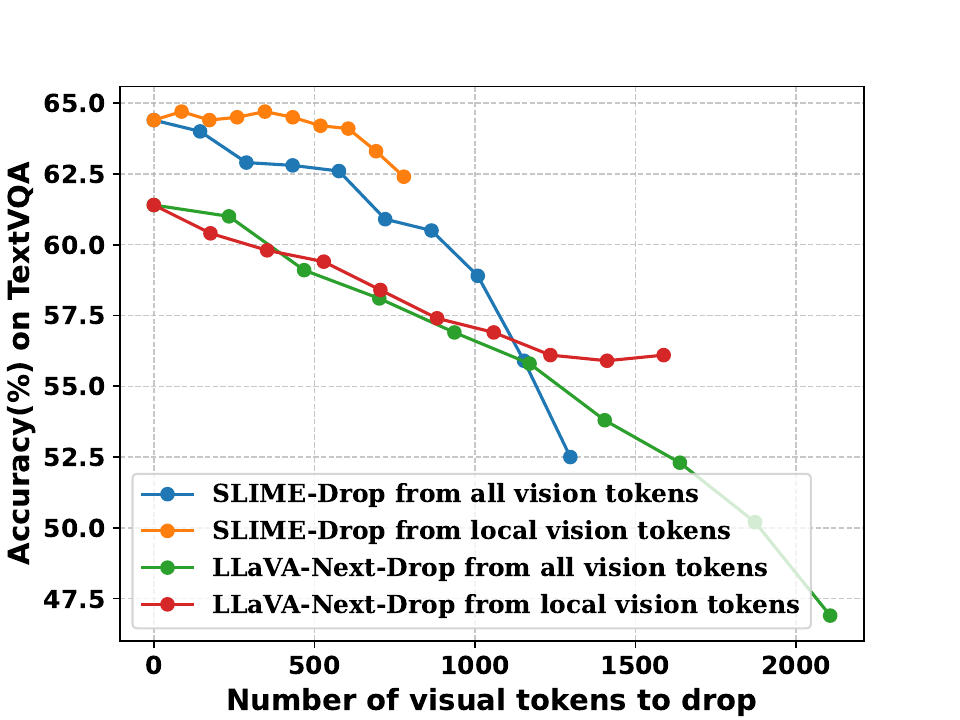}
        \caption{Performance under different dropping numbers on TextVQA.}
        \label{fig:sparsity_analysis_textvqa}
    \end{subfigure}
    \hfill
    \begin{subfigure}[b]{0.325\textwidth}
        \centering
        \includegraphics[width=\textwidth]{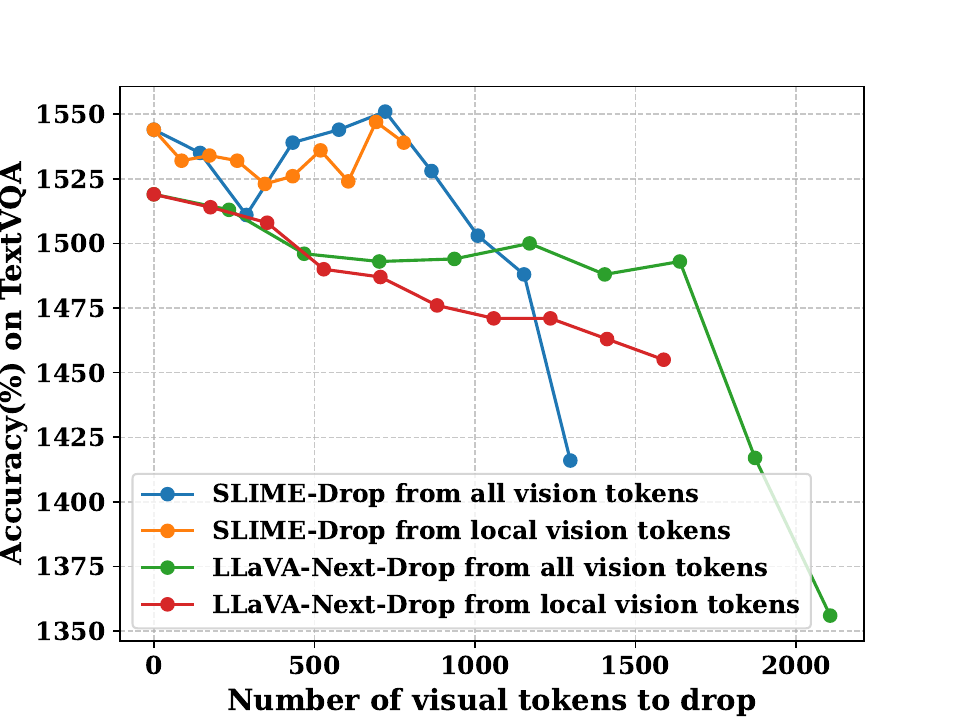}
        \caption{Performance under different dropping numbers on MME.}
        \label{fig:sparsity_analysis_mme}
    \end{subfigure}
    \hfill
    \begin{subfigure}[b]{0.325\textwidth}
        \centering
        \includegraphics[width=\textwidth]{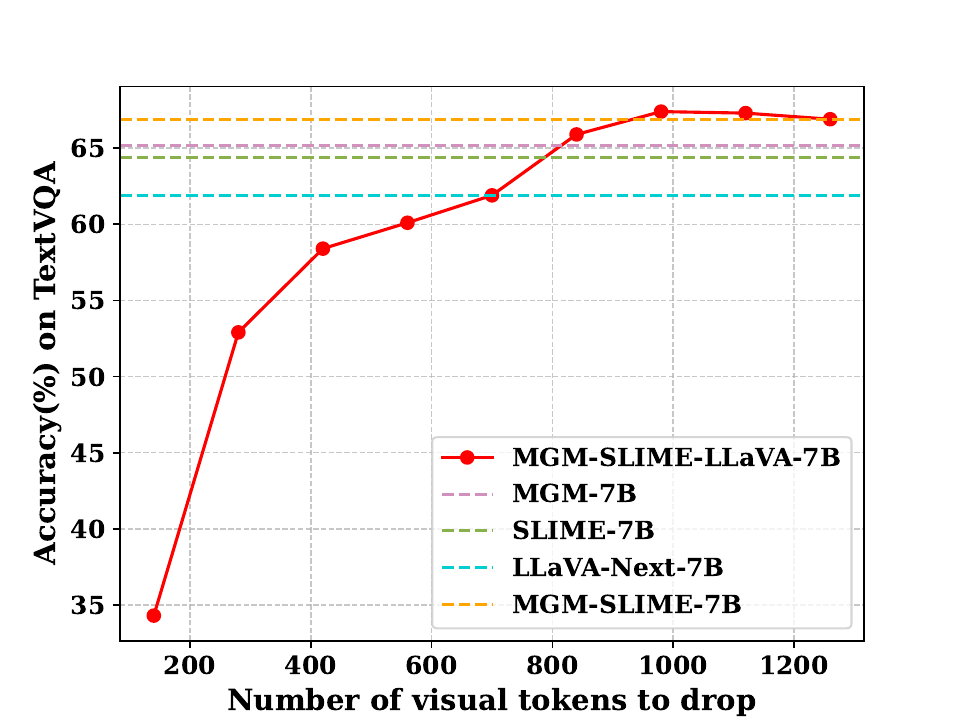}
        \vspace{-0.5cm}
        \caption{Performance under different dropping numbers of integration.}
        \label{fig:pruning_3models_textvqa}
    \end{subfigure}
    \caption{Exploration of redundancy in visual tokens: (a) and (b) examine the redundancy of visual tokens in SLIME-7B and LLaVA-Next-7B across the TextVQA and MME datasets. (c) investigates the effect of randomly dropping tokens on performance after integrating these two models.}
    \label{fig:appendix_c2}
\end{figure*}

To address the issue of inconsistency between training and testing caused by excessively long input sequences, one straightforward solution is to fine-tune the model with longer visual input sequences. As shown in Table~\ref{tab:main_results_appendix}, the 8B models, having been trained on longer token sequences, demonstrate significant performance gains with the direct integration of the three MLLMs. However, training on longer sequences comes with considerable computational costs. To mitigate this challenge more efficiently, we investigate the possibility of directly reducing the token sequence length before LLM's inference. Previous work on token pruning suggests that visual features in MLLMs exhibit substantial redundancy~\citep{bolya2022tome,chen2024fastv}, indicating that pruning redundant tokens could be a feasible solution.
In the following, we assess the effectiveness of token pruning in mitigating the negative impact of inconsistent sequence lengths between training and testing, using the integration of MGM-7B, LLaVA-Next-7B, and SLIME-7B as case studies.

\begin{table*}[htp]
\caption{Performance of integrating more than two MLLMs. Res. indicates the resolution of the input image. Percentages indicate the rate of improvement compared to the best performance of the baseline. * denotes randomly dropping 1000 vision tokens.}
\label{tab:main_results_appendix}
\centering
\resizebox{0.95\textwidth}{!}{
\begin{tabular}{l|l|c|c|c|c|c|c|c}
\toprule
\textbf{Method} & \textbf{LLM} & \textbf{Res.} & \textbf{VQA\(^T\)} & \textbf{MME\(^P\)} & \textbf{MME\(^E\)} & \textbf{MMB} & \textbf{MMMU$_v$} & \textbf{MMMU$_t$} \\
\midrule
MGM-7B & Vicuna-7B & 336 & 65.2 & 1523 & 316 & 68.7 & 36.1 & 32.8 \\ 
SliME-7B & Vicuna-7B & 336 & 64.4 & 1544 & 383 & 68.4 & 37.2 & 33.4 \\ 
LLaVA-Next-7B & Vicuna-7B & 336 & 61.9 & 1519 & 334 & 65.6 & 30.7 & 30.5 \\ 
\rowcolor{gray!20}
MGM-SliME-7B & Vicuna-7B & 336 & \textbf{66.9} (\textcolor{blue}{+2.6\%}) & \textbf{1563} (\textcolor{blue}{+1.2\%}) & \textbf{394} (\textcolor{blue}{+2.8\%}) & \textbf{69.6} (\textcolor{blue}{+1.3\%}) & \textbf{37.8} (\textcolor{blue}{+1.6\%}) & \textbf{33.6} (\textcolor{blue}{+0.6\%}) \\ 
\rowcolor{gray!20}
MGM-SliME-LLaVA-7B & Vicuna-7B & 336 & 20.1 (\textcolor{red}{-69.2\%}) & 1220 (\textcolor{red}{-21.0\%}) & 268 (\textcolor{red}{-30.2\%}) & 60.7 (\textcolor{red}{-13.2\%}) & 29.9 (\textcolor{red}{-20.9\%}) & 21.2 (\textcolor{red}{-38.0\%}) \\ 
\rowcolor{gray!20}
MGM-SliME-LLaVA-7B$^*$ & Vicuna-7B & 336 & \textbf{67.4} (\textcolor{blue}{+3.4\%}) & \textbf{1570} (\textcolor{blue}{+1.7\%}) & \textbf{397} (\textcolor{blue}{+3.7\%}) & \textbf{70.2} (\textcolor{blue}{+2.2\%}) & \textbf{37.8} (\textcolor{blue}{+1.6\%}) & \textbf{33.9} (\textcolor{blue}{+1.5\%}) \\ 
\midrule
MGM-8B & LLaMA-3-8B-Instruct & 336 & 67.6 & 1606 & 341 & 68.1 & 38.2 & 36.3 \\ 
SliME-8B & LLaMA-3-8B-Instruct & 336 & 64.8 & 1578 & 337 & 73.2 & 40.8 & 37.2 \\ 
LLaVA-Next-8B & LLaMA-3-8B-Instruct & 336 & 64.6 & 1604 & 318 & 72.1 & 40.7 & 37.0 \\ 
\rowcolor{gray!20}
MGM-SliME-8B & LLaMA-3-8B-Instruct & 336 & \textbf{70.0} (\textcolor{blue}{+3.6\%}) & \textbf{1645} (\textcolor{blue}{+2.4\%}) & \textbf{372} (\textcolor{blue}{+8.3\%}) & \textbf{73.9} (\textcolor{blue}{+1.0\%}) & \textbf{41.6} (\textcolor{blue}{+2.0\%}) & \textbf{38.4} (\textcolor{blue}{+3.2\%}) \\ 
\rowcolor{gray!20}
MGM-SliME-LLaVA-8B & LLaMA-3-8B-Instruct & 336 & \textbf{70.9} (\textcolor{blue}{+4.9\%}) & \textbf{1660} (\textcolor{blue}{+3.4\%}) & \textbf{369} (\textcolor{blue}{+8.2\%}) & \textbf{75.1} (\textcolor{blue}{+2.6\%}) & \textbf{41.7} (\textcolor{blue}{+2.2\%}) & \textbf{38.7} (\textcolor{blue}{+4.0\%}) \\ 
\bottomrule
\end{tabular}
}
\end{table*}

It is worth noting that both SLIME-7B and LLaVA-Next-7B enhance visual representations by incorporating a large number of local features through additional augmentations, resulting in significantly longer visual sequences compared to MGM-7B. We therefore assess the impact of randomly dropping visual tokens on the performance of these two models and explore the extent of redundancy in their local features. Specifically, we test two strategies on the TextVQA and MME datasets: (1) randomly dropping a specified number of tokens from the entire set of visual tokens, and (2) randomly dropping a specified number of tokens exclusively from the local features. As shown in Figures~\ref{fig:sparsity_analysis_textvqa} and \ref{fig:sparsity_analysis_mme}, randomly dropping SLIME-7B’s visual tokens leads to a substantial performance decrease, whereas removing fewer than 500 visual tokens exclusively from the local features does not result in a significant performance loss. This indicates that SLIME-7B’s local features exhibit considerable redundancy. In contrast, random token pruning from either the full token set or just the local features in LLaVA-Next-7B results in a notable performance decline. Based on these observations, we recommend pruning more visual tokens in SLIME-7B while limiting the number of tokens drop in LLaVA-Next-7B.

We implement a strategy of randomly dropping visual tokens to reduce the token sequence length, enabling the language model to accommodate a larger number of visual tokens from different models. We first conduct experiments on the TextVQA dataset by integrating the MGM-7B, SLIME-7B, and LLaVA-Next-7B models. As shown in Figure~\ref{fig:pruning_3models_textvqa}, performance improves when a certain number of tokens are dropped, after which further pruning may lead to a decline in performance. Ultimately, when approximately 1000 tokens are dropped, the performance of the three-model integration surpasses that of the two-model integration. Additionally, as presented in Table~\ref{tab:main_results_appendix}, random pruning of 1000 tokens leads to a notable improvement in overall performance. However, different MLLMs exhibit varying degrees of token sparsity, making it inconvenient to test each one before integration. We leave the design of a more efficient cross-model token fusion/pruning strategy for future work.

\section{Discussion of Efficiency}
\label{sec:efficiency}
As mentioned in Section 4, the increase in FLOPs for~\methodname~primarily arises from the increased length of visual tokens. 
To mitigate the significant rise in inference cost caused by extended visual sequences, a simple approach is to prune the token sequence. FastV~\citep{chen2024fastv} identifies substantial redundancy in visual tokens starting from the $3^{rd}$ layer. Inspired by this, we use the integration of MGM-8B and SLIME-8B as a case study to explore the extent of redundancy in visual tokens after integrating multiple MLLMs, and we compare the FLOPs and inference time. As shown in Table~\ref{tab:flops_and_time}, after pruning 50\% vision tokens starting from the $3^{rd}$ layer, the FLOPs of~\methodname~are reduced to levels below those of both MGM-8B and SLIME-8B, while maintaining superior performance compared to the individual models.

\vspace{-0.2cm}
\begin{table*}[ht]
\caption{The comparison of FLOPs and inference time under different token pruning ratios. The results of~\methodname~are marked in gray. We report the average FLOPs and inference time (second) per sample on the TextVQA and MME datasets. ``Sparsity" refers to the pruning rate of vision tokens, where 0\% indicates no sparsity. ``Params" indicates the number of parameters. We evaluate the inference time on a single NVIDIA A800 GPU.}
\label{tab:flops_and_time}
\centering
\resizebox{0.95\textwidth}{!}{
\begin{tabular}{l|c|c|ccc|ccc}
\toprule
\textbf{Method} & \textbf{Sparsity} & \textbf{Res.} & \textbf{TextVQA} & \textbf{MME\(^P\)} & \textbf{MME\(^E\)} & \textbf{Params} & \textbf{FLOPs} & \textbf{Inference Time}   \\
\midrule
MGM-8B & 0\% & 336 & 67.6 & 1606 & 341 & 8.6B & 22.73T & 0.2395 \\ 
MGM-8B & 30\% & 336 & 66.5 & 1536 & 306 & 8.6B & 9.14T & 0.2354 \\ 
MGM-8B & 50\% & 336 & 65.5 & 1529 & 316 & 8.6B & 7.54T & 0.2292 \\ 
\midrule
SliME-8B & 0\% & 336 & 64.8 & 1578 & 337 & 8.4B & 79.04T & 0.3586 \\ 
SliME-8B & 30\% & 336 & 64.4 & 1584 & 337 & 8.4B & 18.71T & 0.3370 \\ 
SliME-8B & 50\% & 336 & 63.4 & 1579 & 335 & 8.4B & 14.86T & 0.2800 \\ 
\midrule
\rowcolor{gray!20}
MGM-SliME-8B & 0\% & 336 & \textbf{70.0} & 1645 & \textbf{372} & 8.9B & 141.56T & 0.4171 \\ 
\rowcolor{gray!20}
MGM-SliME-8B & 30\% & 336 & 69.8 & \textbf{1649} & 368 & 8.9B & 27.18T & 0.4080 \\ 
\rowcolor{gray!20}
MGM-SliME-8B & 50\% & 336 & 69.8 & 1637 & 370 & 8.9B & 21.54T & 0.3624 \\
\rowcolor{gray!20}
MGM-SliME-8B & 70\% & 336 & 68.5 & 1597 & 351 & 8.9B & 16.03T & 0.3359 \\ 
\bottomrule
\end{tabular}
}
\end{table*}

\section{Performance on Integration of MLLMs from Different Families}
To further evaluate the performance of~\methodname~on MLLMs from different families, we conduct experiments with MGM-7B, MGM-8B~\citep{li2024mini}, VILA-7B, and VILA-8B~\citep{lin2024vila}, as shown in Table~\ref{tab:different_families}. Vicuna-v1.5 is trained on LLaMA-2-7B, and LLaMA-3-8B-Instruct shares the same architecture as LLaMA-3-8B, meaning they have the same structure but different parameters. For merging the language model of MGM-7B and VILA-7B, we compute the delta parameters based on LLaMA-2-7B, and for merging the language model of VILA-8B and MGM-8B, we use delta parameters based on LLaMA-3-8B. Notably, the delta parameters differences between MGM-7B and VILA-7B are larger compared to those between MGM-7B and SLIME-7B, and similarly, the delta parameters differences between MGM-8B and VILA-8B are larger compared to those between MGM-8B and SLIME-8B. While improvements are achieved on some datasets, there is a significant performance drop on others. This is attributed to the substantial differences in delta parameters, resulting in a significant alignment error between the merged language model and the individual vision encoders.

\begin{table*}[ht]
\caption{Merging multimodal large language models from different families.}
\label{tab:different_families}
\centering
\resizebox{0.95\textwidth}{!}{
\begin{tabular}{l|c|ccccccc}
\toprule
\textbf{Method} & \textbf{LLM}  & \textbf{TextVQA} & \textbf{MME\(^P\)} & \textbf{MME\(^E\)} & \textbf{GQA}  & \textbf{POPE} & \textbf{MMMU$_v$}  \\
\midrule
MGM-7B & Vicuna-v1.5 & 65.2 & 1523 & \textbf{316} & \textbf{64.5} & 84.1 & \textbf{36.1} \\
VILA-7B & LLaMA-2-7B  & 64.4 & 1533 & 293 & 62.3 & \textbf{85.5} & 35.2 \\
\rowcolor{gray!20}
MGM-VILA-7B & LLaMA-2-7B & \textbf{68.6} (\textcolor{blue}{+5.2\%}) & \textbf{1534} (\textcolor{blue}{+0.0\%}) & 298 (\textcolor{red}{-5.7\%}) & 63.4 (\textcolor{red}{-1.7\%}) & 84.9 (\textcolor{red}{-0.7\%}) & 35.3 (\textcolor{red}{-2.2\%}) \\
\midrule
MGM-8B & LLaMA-3-8B-Instruct & \textbf{67.6} & \textbf{1606} & 341 & \textbf{64.3} & \textbf{85.8} & \textbf{38.2} \\ 
VILA-8B & LLaMA-3-8B & 66.3 & 1577 & 326 & 61.9 & 84.4 & 36.9 \\
\rowcolor{gray!20}
MGM-VILA-8B & LLaMA-3-8B & 33.0 (\textcolor{red}{-51.2\%}) & 1496 (\textcolor{red}{-6.8\%}) & \textbf{349} (\textcolor{blue}{+2.3\%}) & 50.3 (\textcolor{red}{-21.8\%}) & 83.9 (\textcolor{red}{-2.2\%}) & 33.2 (\textcolor{red}{-13.1\%})   \\
\bottomrule
\end{tabular}
}
\end{table*}

\section{More Visualizations}
We provide additional samples and visualizations to further demonstrate the effectiveness of our method. We visualize the average cross-attention across all layers of the model before and after integration. `MGM' refers to MGM-8B, and `SLIME' refers to SLIME-8B.

\begin{figure*}[tb]
    \centering
    
    \begin{subfigure}[b]{0.468\textwidth}
        \centering
        \includegraphics[width=\textwidth]{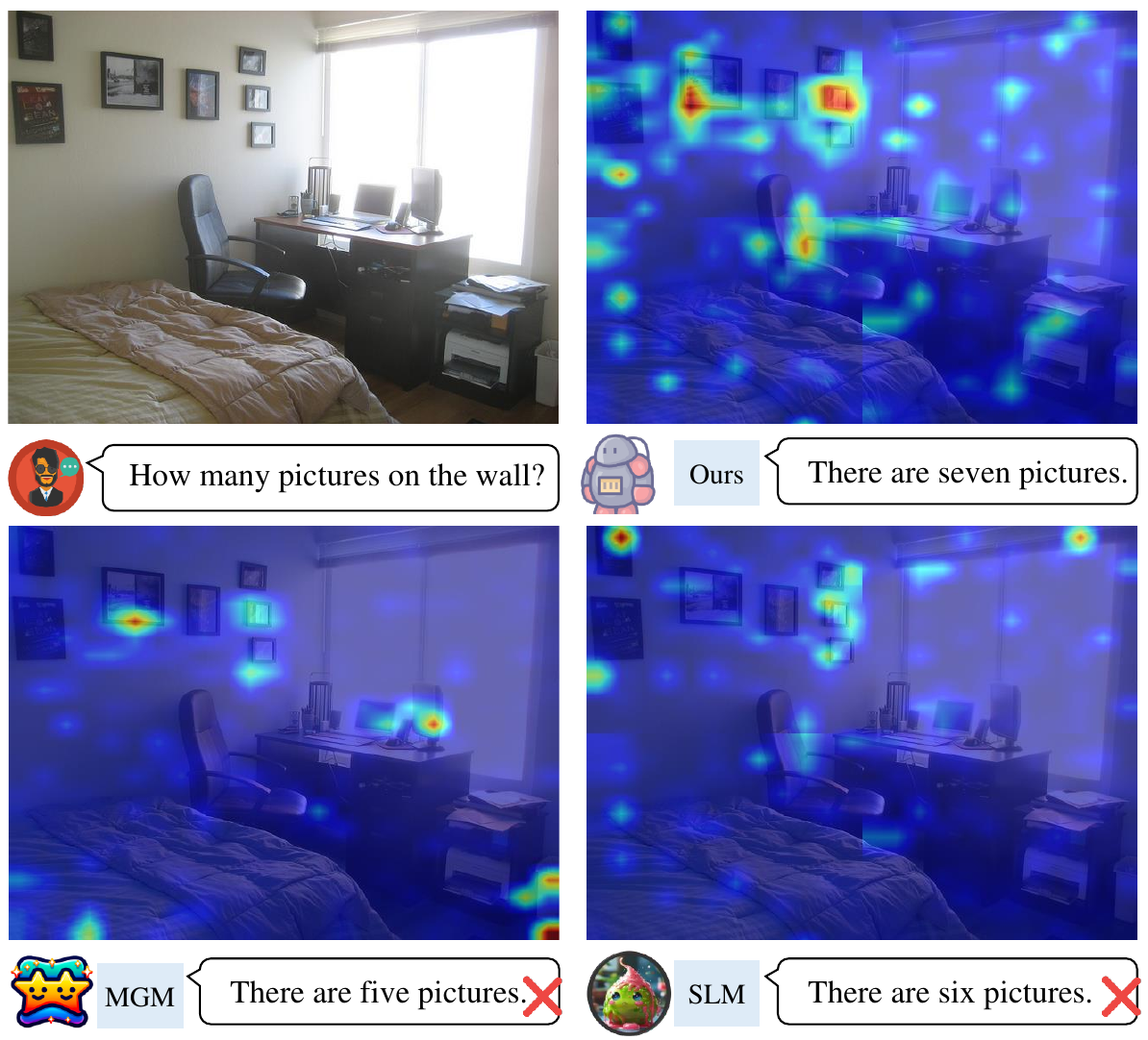}
        \label{fig:vis_cross_1}
    \end{subfigure}
    \hfill
    \begin{subfigure}[b]{0.522\textwidth}
        \centering
        \includegraphics[width=\textwidth]{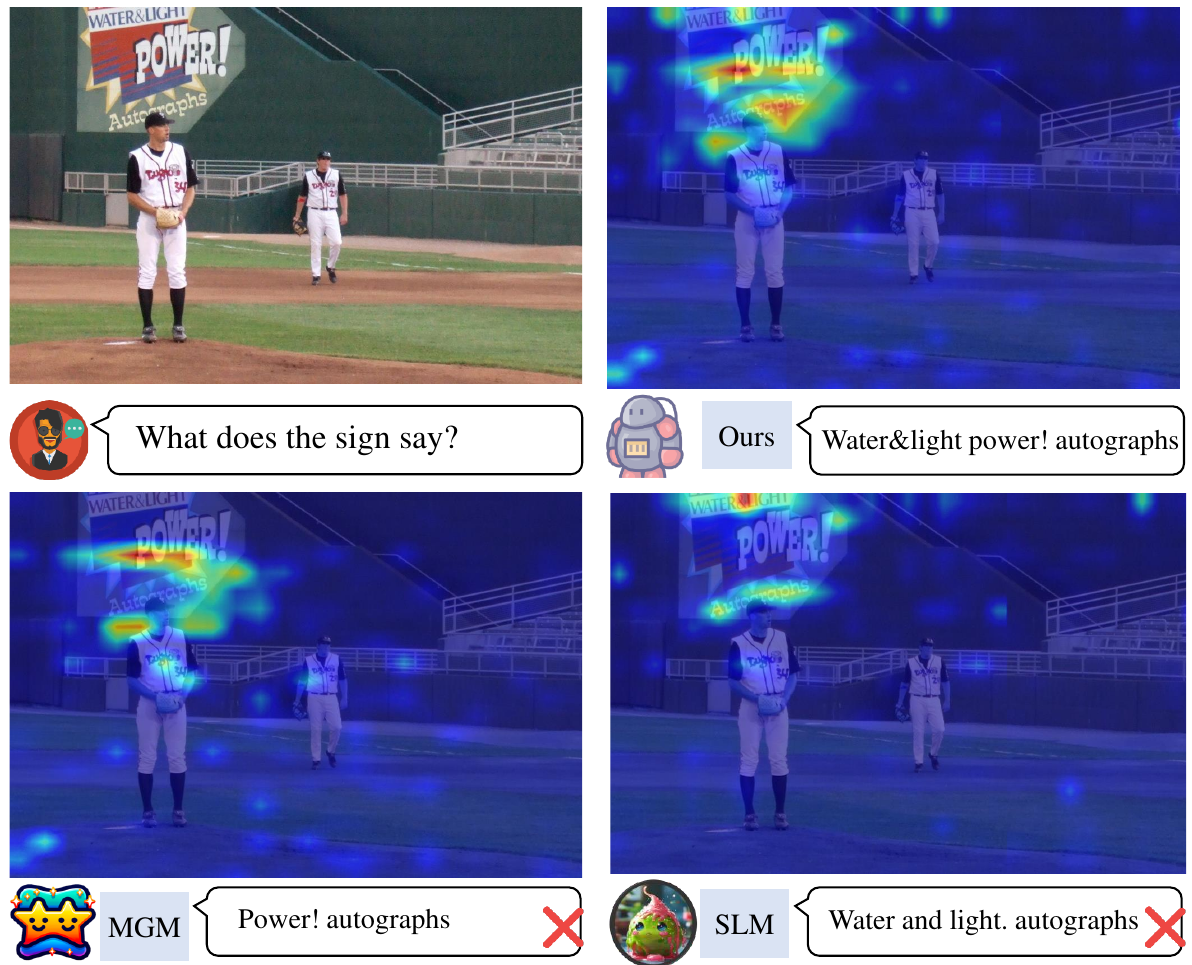}
        \label{fig:image2}
    \end{subfigure}
    \caption{Enhancement of the perception regions.}
    \label{fig:vis_two_cross_attn}
\end{figure*}

In Figure~\ref{fig:vis_two_cross_attn}, it can be observed that after integrating multiple MLLMs using our method, the model captures more information and attends to a wider range of target regions. For example, in the left sample, the task is to count the number of pictures on the wall. Each model fails to detect all the pictures, resulting in an incorrect answer. However, after integrating multiple MLLMs using \methodname, the model successfully attends to all the pictures and provides an accurate answer. In the right sample, individual models tend to focus on only a portion of the sign, leading to incomplete answers. In contrast, \methodname~enables the model to attend to a more comprehensive region, resulting in a more accurate response.

\begin{figure*}[ht]
    \centering
    \begin{subfigure}[b]{0.49\textwidth}
        \centering
        \includegraphics[width=\textwidth]{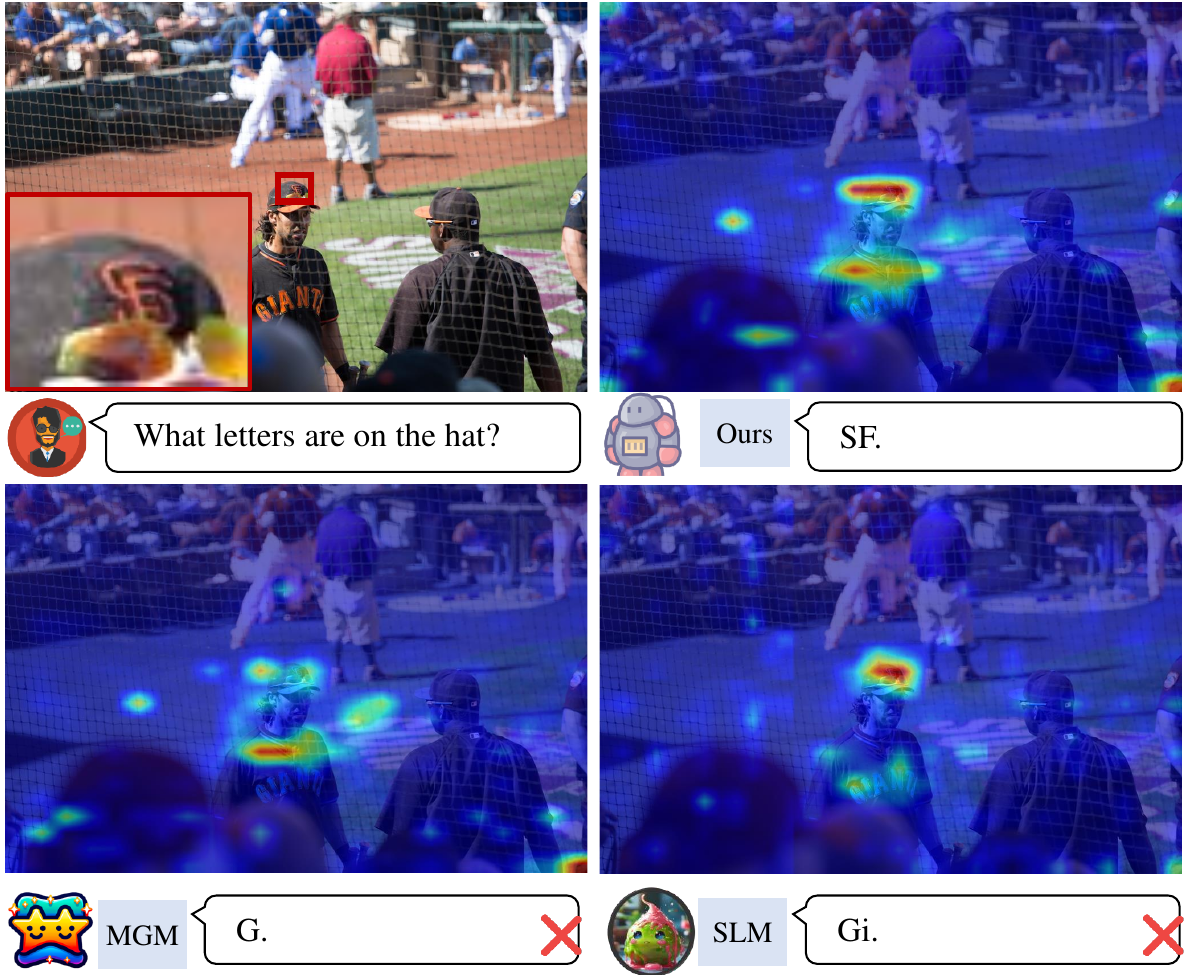}
        \label{fig:vis_textvqa_1}
    \end{subfigure}
    \hfill
    \begin{subfigure}[b]{0.49\textwidth}
        \centering
        \includegraphics[width=\textwidth]{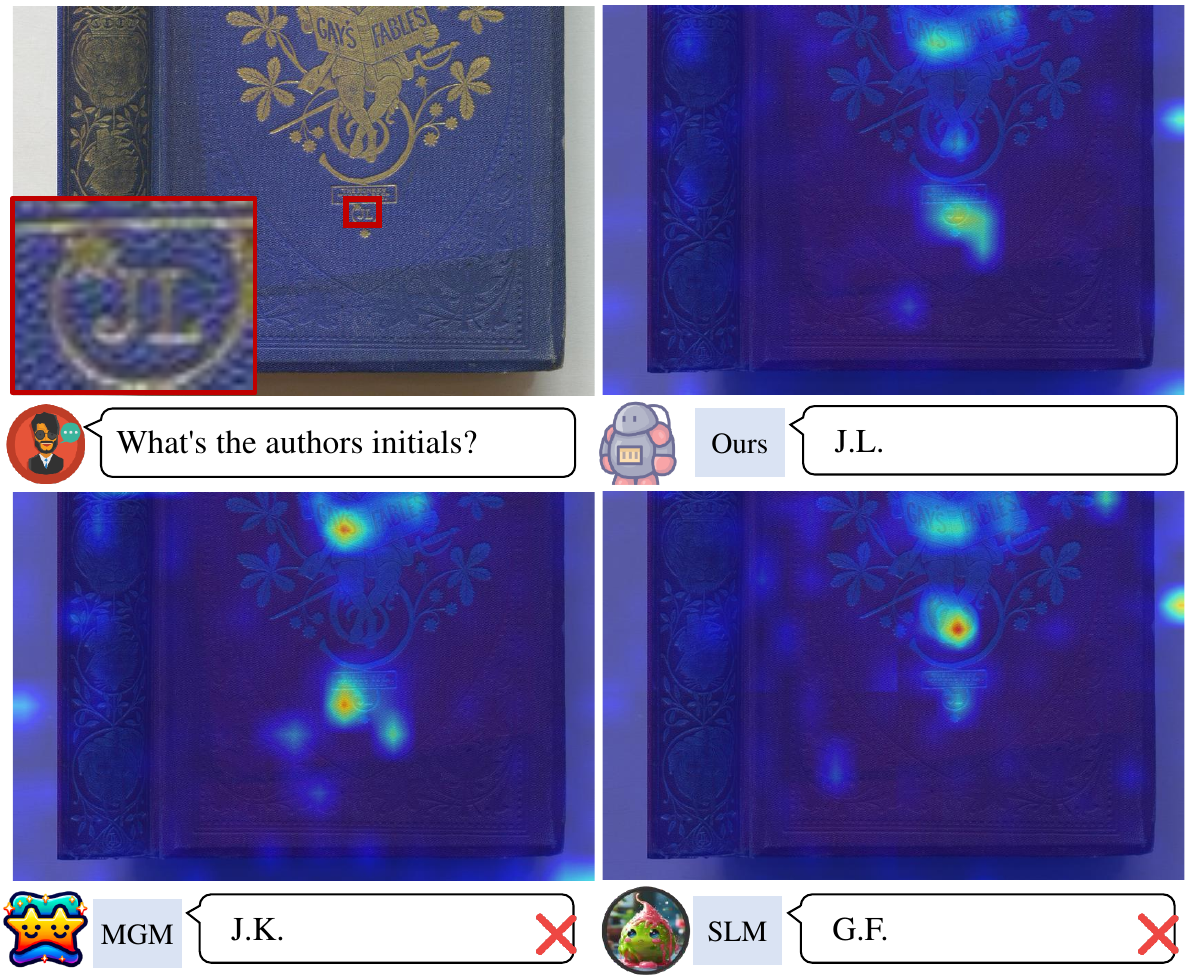}
        \label{fig:vis_textvqa_2}
    \end{subfigure}
    \caption{Enhancement of fine-grained perception}
    \label{fig:vis_textvqa}
\end{figure*}

In Figure~\ref{fig:vis_textvqa}, we visualize two examples from the TextVQA dataset. The target texts in these examples are extremely small, requiring the model to possess strong fine-grained perception capabilities for accurate recognition. While both MGM-8B and SLIME-8B identify the target regions, neither is able to provide the correct answers. In contrast, our method, which concatenates visual tokens from both models, significantly enhances the model's fine-grained perception, enabling it to correctly identify the results.

\begin{figure*}[ht]
    \centering
    \begin{subfigure}[b]{0.49\textwidth}
        \centering
        \includegraphics[width=\textwidth]{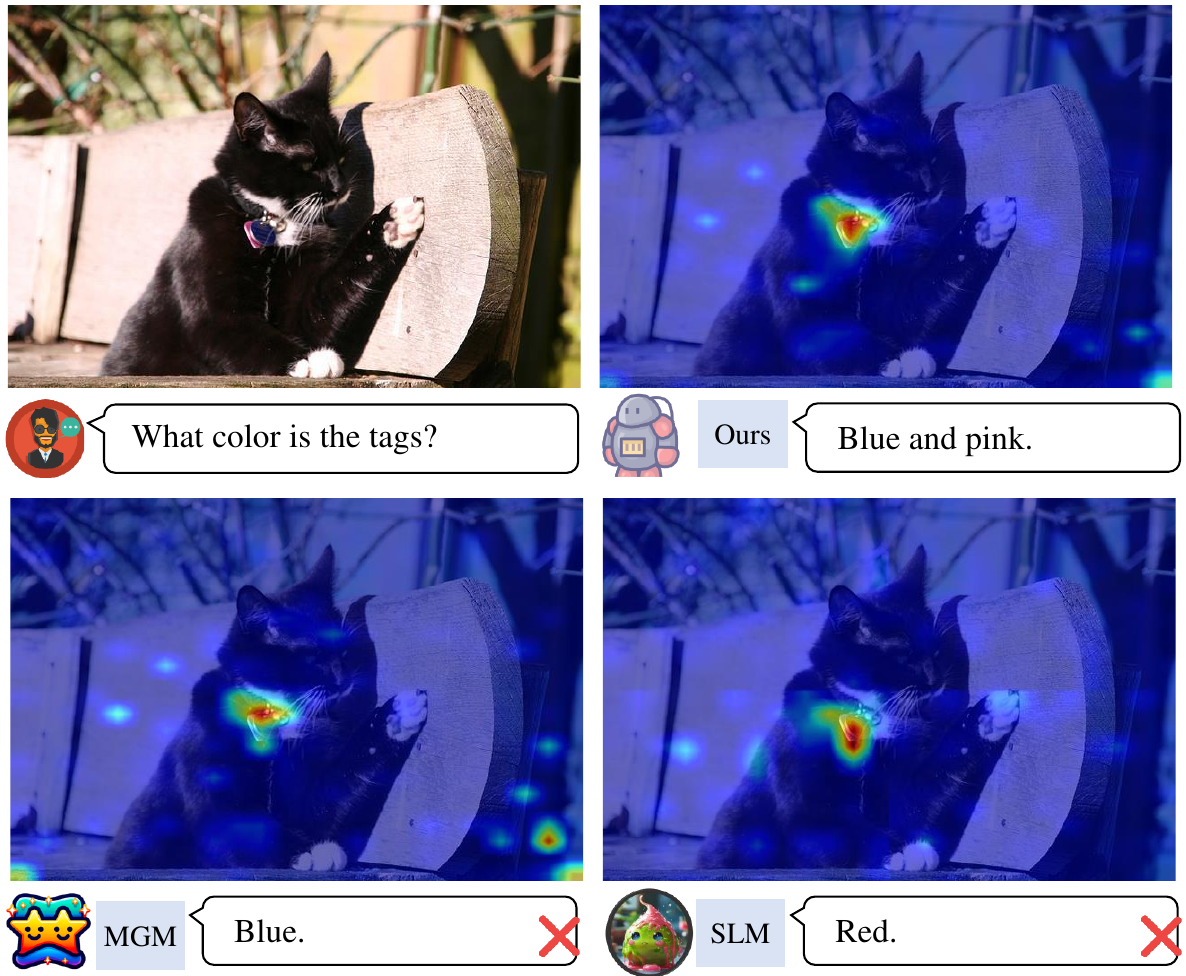}
        \label{fig:vis_vqav2_1}
    \end{subfigure}
    \hfill
    \begin{subfigure}[b]{0.49\textwidth}
        \centering
        \includegraphics[width=\textwidth]{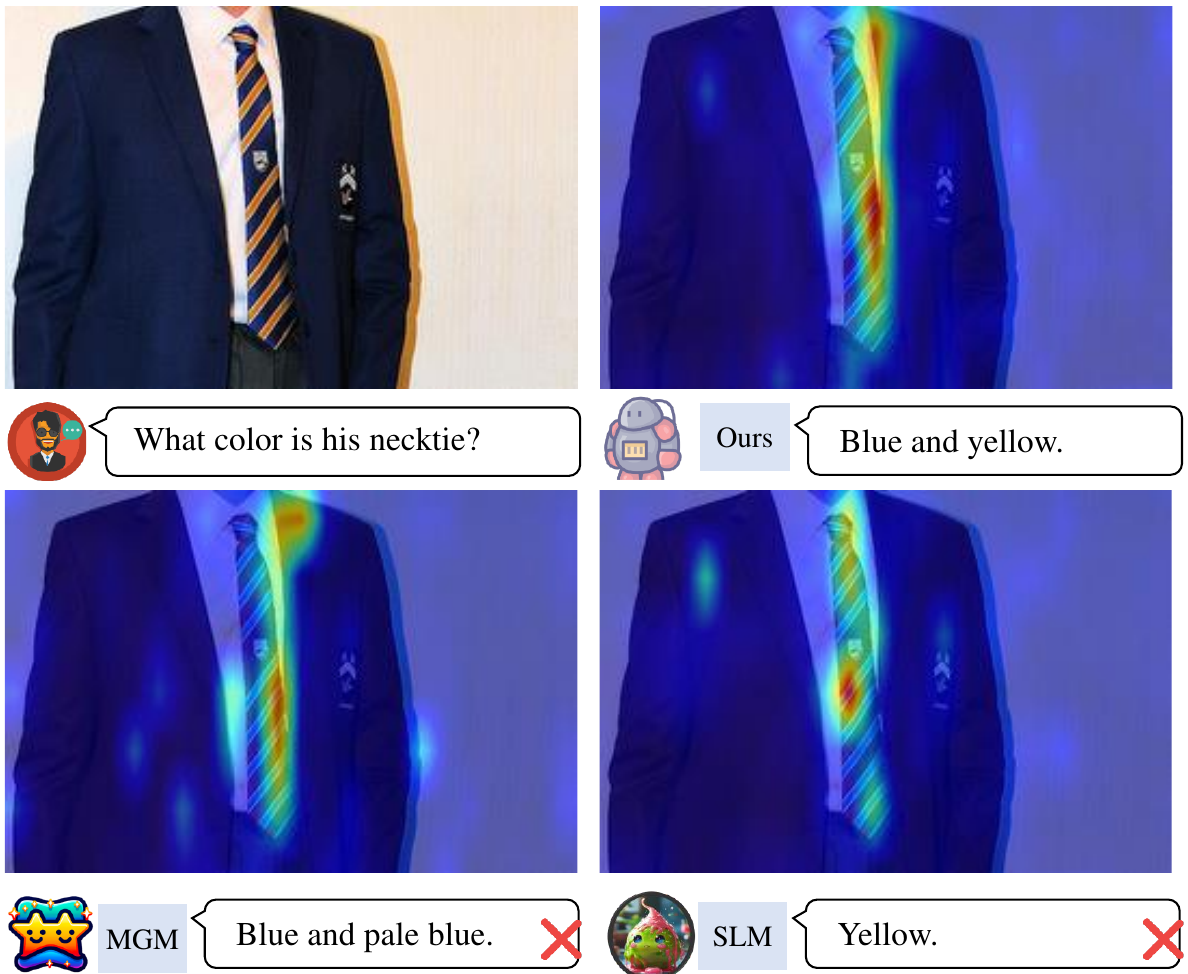}
        \label{fig:vis_vqav2_2}
    \end{subfigure}
        \caption{Enhancement of colors perception}
    \label{fig:vis_vqav2}
\end{figure*}

In Figure~\ref{fig:vis_vqav2}, we present two examples from the VQAv2 dataset. Although both MGM-8B and SLIME-8B successfully attend to the target regions, neither could correctly identify the color of the target object. Interestingly, the colors perceived by the two models are complementary to the actual target colors. After integrating the two models, our method is able to accurately recognize the correct color.

\begin{figure*}[ht!]
    \centering
    \begin{subfigure}[b]{0.99\textwidth}
        \centering        \includegraphics[width=\textwidth]{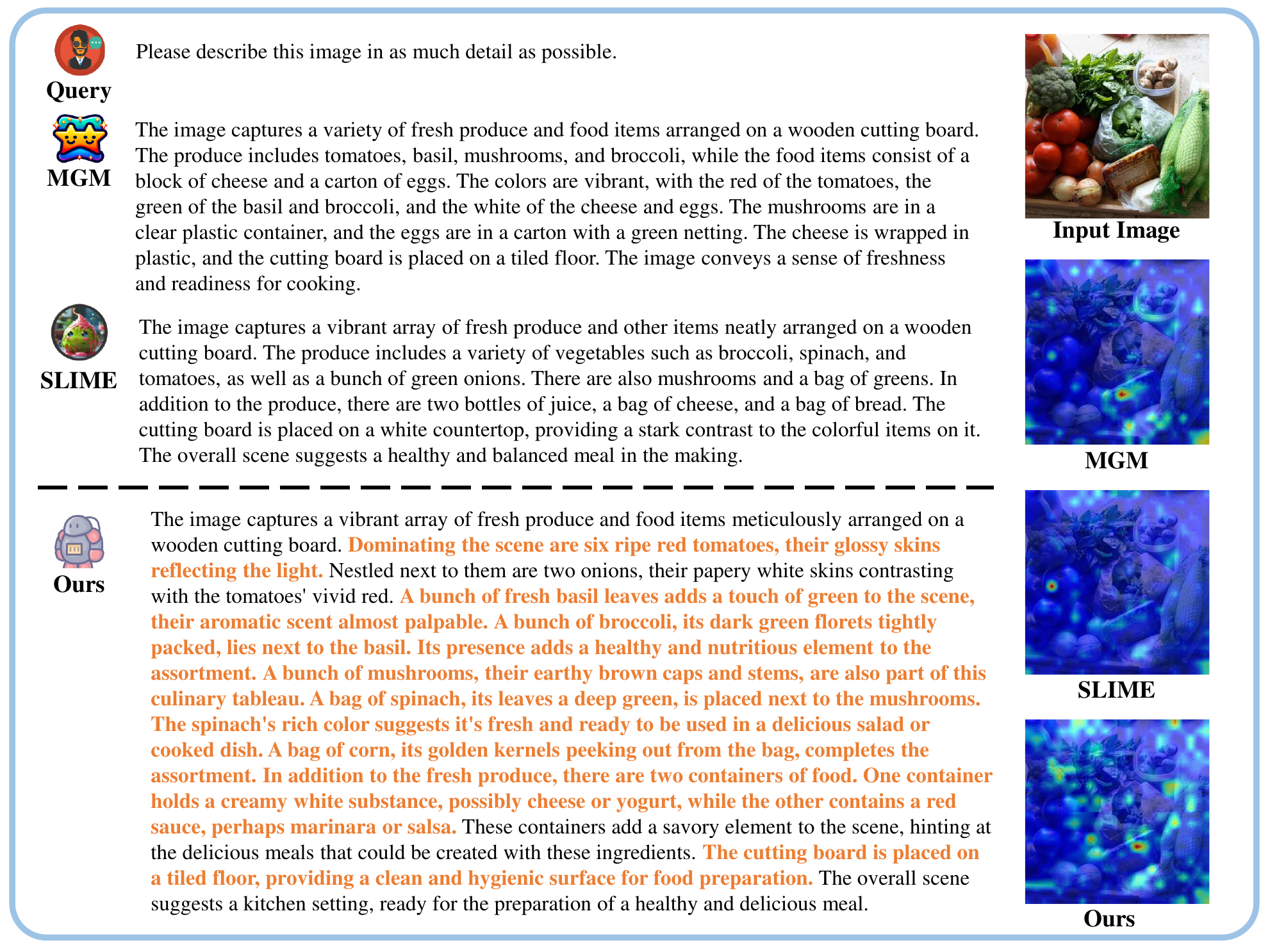}
    \end{subfigure}
    \caption{A sample for detailed descriptions.}
    \label{fig:vis_gpt_1}
\end{figure*}

In Figure~\ref{fig:vis_gpt_1}, our method captures more comprehensive image details compared to individual models. For instance, MGM mentions onions and the location of the cutting board, while SLIME does not. Conversely, SLIME identifies spinach, which MGM overlooks. The unique visual information captured by these individual models is effectively combined when integrated using our method.

\begin{figure*}[ht!]
    \centering
    \begin{subfigure}[b]{0.995\textwidth}
        \centering
        \includegraphics[width=\textwidth]{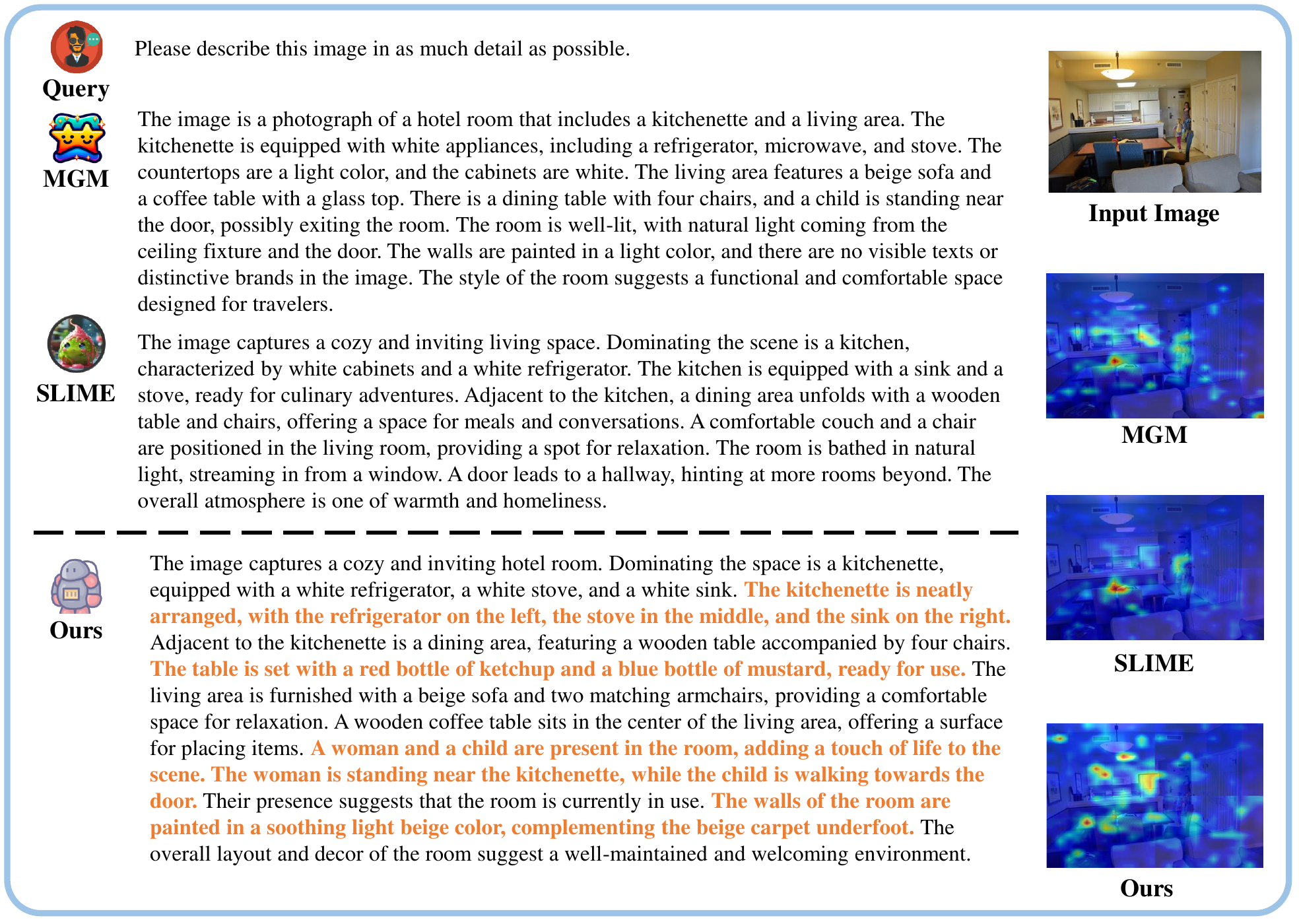}
    \end{subfigure}
        \caption{A sample for detailed descriptions.}
        \label{fig:vis_gpt_appendix1}
\end{figure*}

In Figure~\ref{fig:vis_gpt_appendix1}, MGM focuses on the overall layout of the room, mentioning the white appliances in the kitchenette and the presence of a child near the door, while SLIME emphasizes the cozy atmosphere of the kitchen and living room, describing the natural light and hinting at additional rooms beyond the door. Our method, however, not only retains the overall layout but also highlights key details, such as the red ketchup bottle, the blue mustard bottle, and the presence of a woman and child, adding life and richness to the scene.

\begin{figure*}[ht!]
    \begin{subfigure}[b]{0.995\textwidth}
        \centering
        \includegraphics[width=\textwidth]{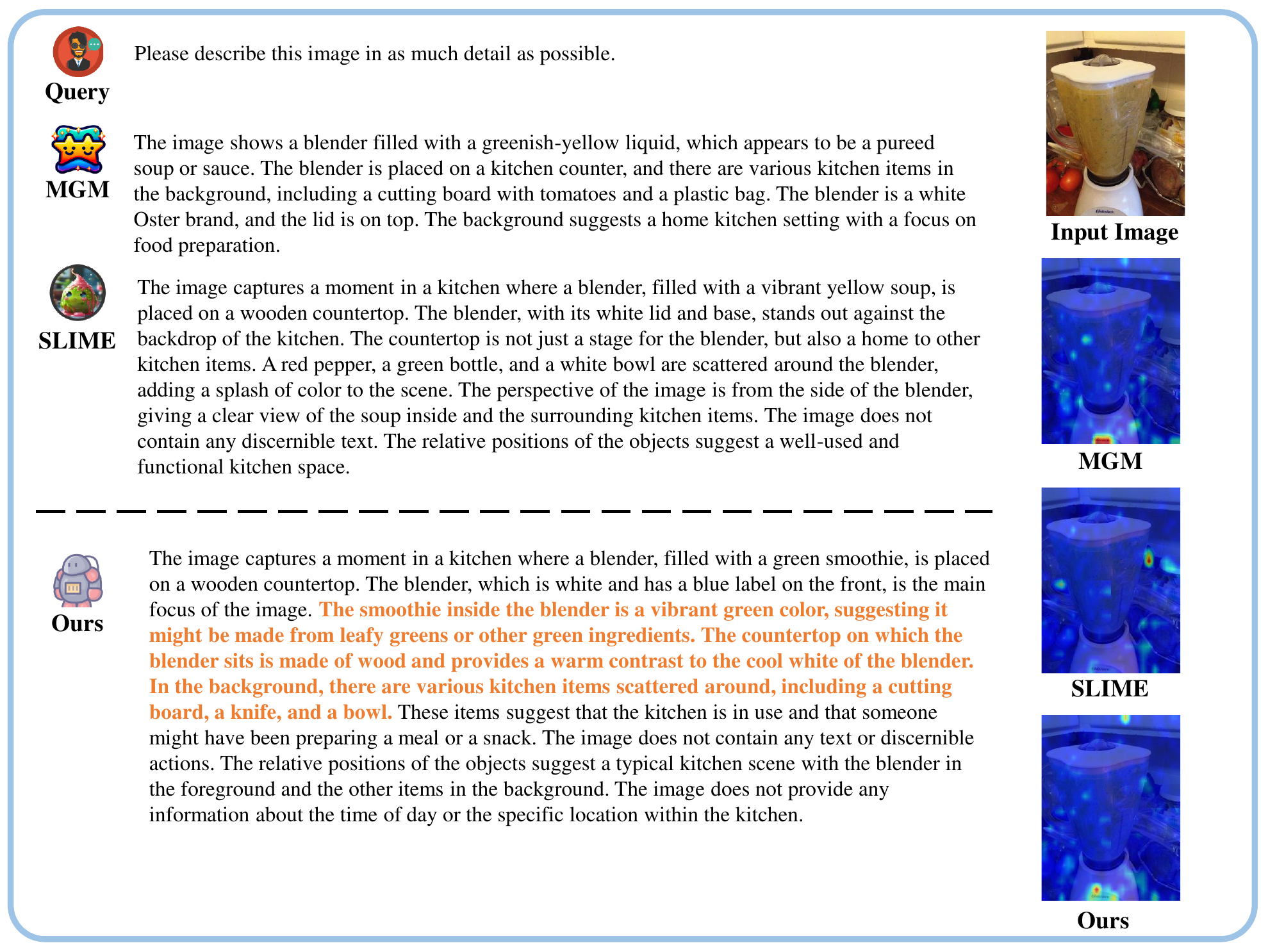}
    \end{subfigure}
        \caption{A sample for detailed descriptions.}
        \label{fig:vis_gpt_appendix2}
    \label{fig:appendix_vis_gpt1}
\end{figure*}

In Figure~\ref{fig:vis_gpt_appendix2}, MGM describes the kitchen, noting the blender, the soup’s greenish-yellow color, and objects like a cutting board and a plastic bag in the background. SLIME, on the other hand, focuses on vibrant details, mentioning objects like a red pepper, a green bottle, and a white bowl. Our method captures all these elements while adding further details, such as the smoothie being green—suggesting it may be made from leafy greens—and the contrast between the wooden countertop and the white blender, enriching the contextual understanding of the scene.

\begin{figure*}[ht!]
    \centering
    \begin{subfigure}[b]{0.995\textwidth}
        \centering
        \includegraphics[width=\textwidth]{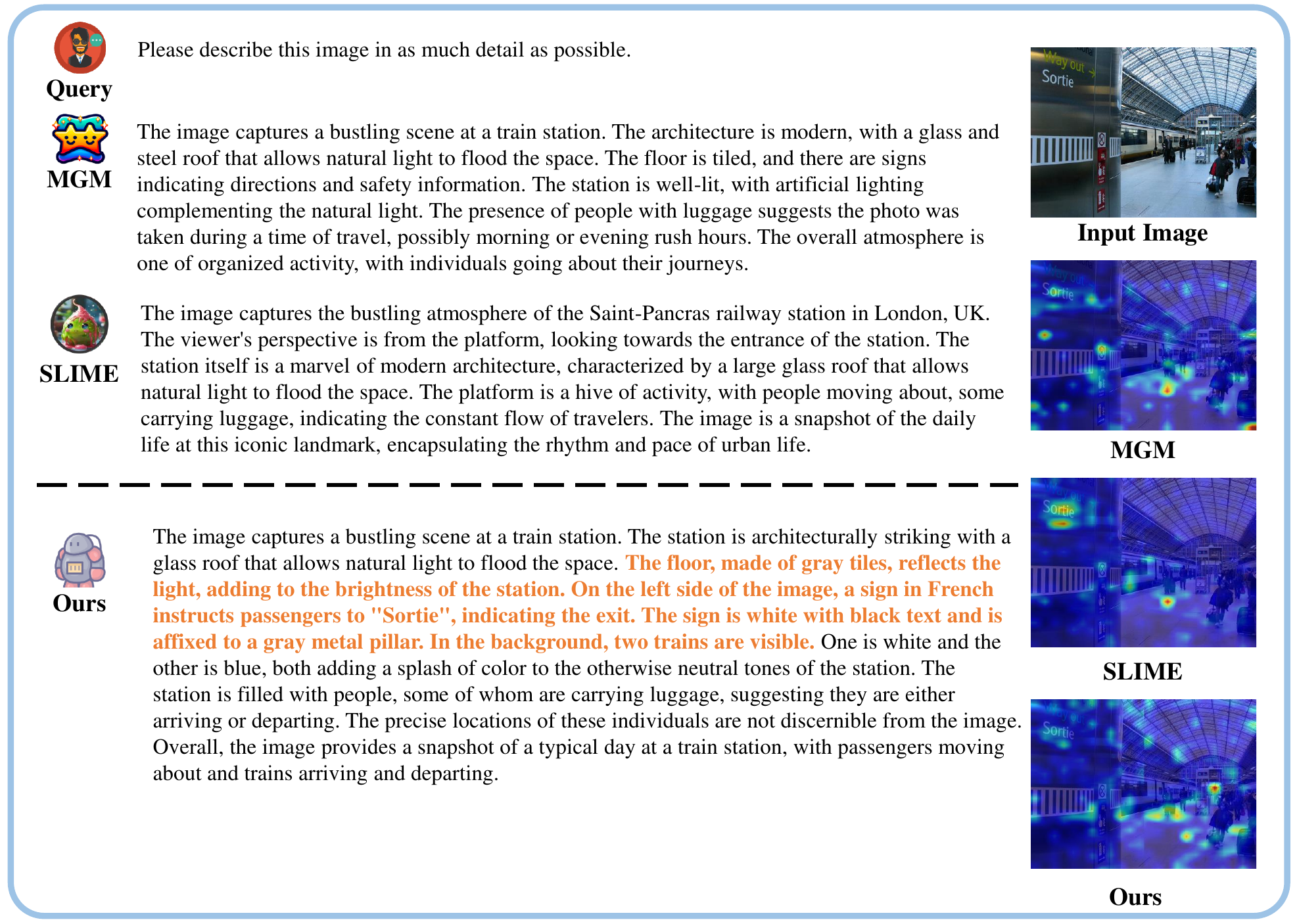}
    \end{subfigure}
        \caption{A sample for detailed descriptions.}
        \label{fig:vis_gpt_appendix3}
\end{figure*}

In Figure~\ref{fig:vis_gpt_appendix3}, MGM highlights the modern architecture of the train station, focusing on the glass roof that lets in natural light and the presence of travelers with luggage, reflecting a busy travel period. SLIME provides more specificity by identifying the station as Saint Pancras in London and emphasizing the constant flow of passengers. Our method combines these insights and adds further details, such as the gray tiles reflecting light, the French sign directing passengers to the exit, and the two visible trains in the background, one white and one blue, adding depth and vibrancy to the depiction of the station.

\begin{figure*}[ht!]
    \begin{subfigure}[b]{0.995\textwidth}
        \centering
        \includegraphics[width=\textwidth]{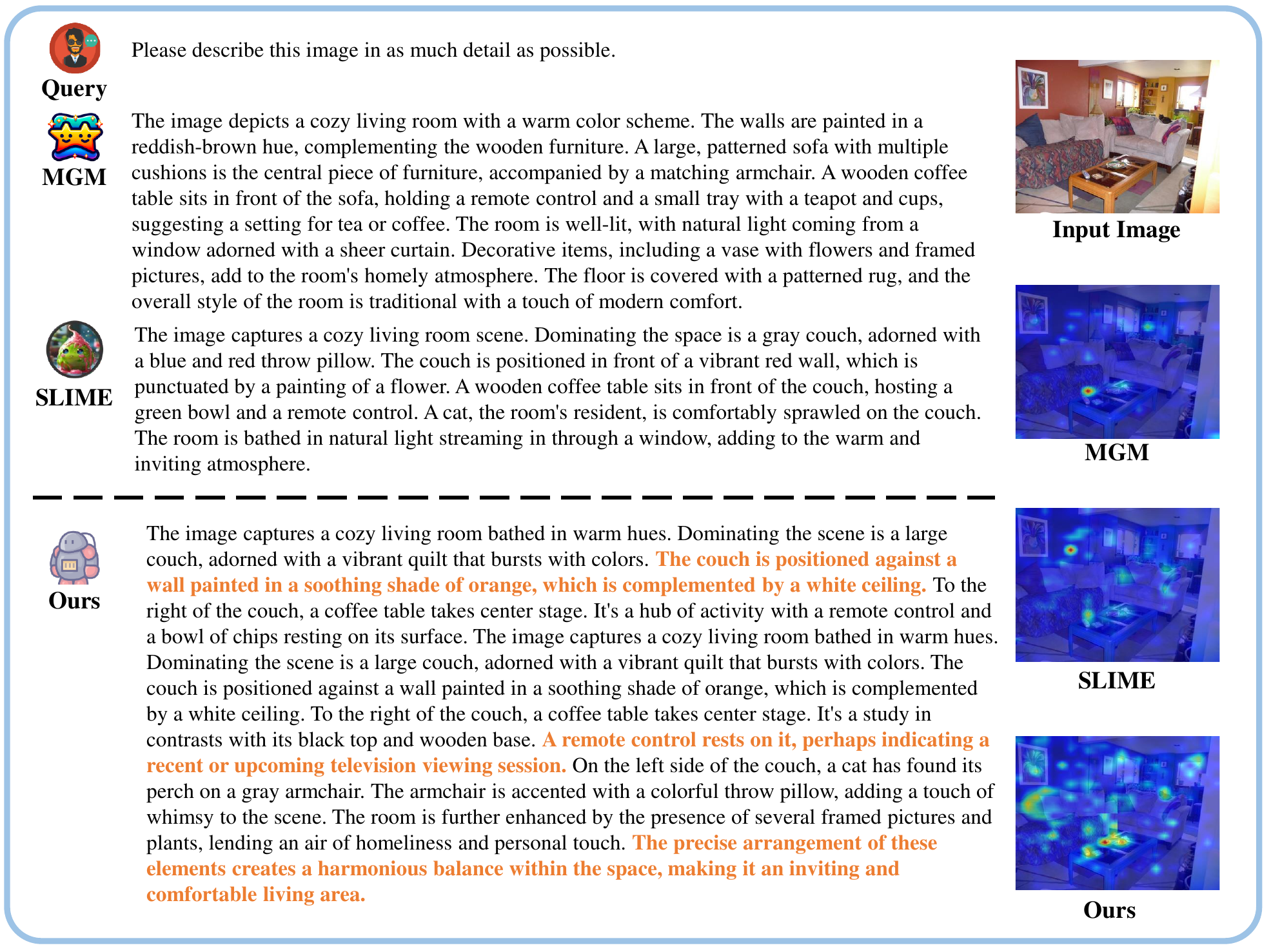}
    \end{subfigure}
        \caption{A sample for detailed descriptions.}
        \label{fig:vis_gpt_appendix4}
    \label{fig:appendix_vis_gpt2}
\end{figure*}

In Figure~\ref{fig:vis_gpt_appendix4}, MGM describes the cozy living room, focusing on the reddish-brown walls, the arrangement of the furniture, and the natural lighting that creates a homely atmosphere. SLIME shifts the focus slightly, noting a gray couch, a vibrant red wall, and a cat lounging on the couch, while also emphasizing the warmth from the natural light. Our method integrates these elements and adds more specific details, such as the orange wall complementing the white ceiling, the remote control on the coffee table, suggesting an upcoming TV session, and the cat perches on the left side of the couch. This combined description provides a more comprehensive and balanced portrayal of the room’s inviting and comfortable ambiance.



\end{document}